\documentclass[preprint,12pt,authoryear]{elsarticle}

\usepackage{algorithm,algpseudocode,float}

\usepackage{lipsum}

\makeatletter
\newenvironment{breakablealgorithm}
  {
   \begin{center}
     \refstepcounter{algorithm}
     \hrule height.8pt depth0pt \kern2pt
     \renewcommand{\caption}[2][\relax]{
       {\raggedright\textbf{\ALG@name~\thealgorithm} ##2\par}
       \ifx\relax##1\relax 
         \addcontentsline{loa}{algorithm}{\protect\numberline{\thealgorithm}##2}
       \else 
         \addcontentsline{loa}{algorithm}{\protect\numberline{\thealgorithm}##1}
       \fi
       \kern2pt\hrule\kern2pt
     }
  }{
     \kern2pt\hrule\relax
   \end{center}
  }
\makeatother

\usepackage{amssymb}
\usepackage{multirow}
\usepackage{lineno}
\modulolinenumbers[5]
\usepackage{afterpage}
\usepackage{color}
\usepackage{makecell}
\usepackage{arydshln} 

\journal{Expert Systems with Applications}

\begin{document}

\begin{frontmatter}



\title{TASL-Net: Tri-Attention Selective Learning Network for Intelligent Diagnosis of Bimodal Ultrasound Video}

\author[mymainaddress]{Chengqian Zhao}
\ead{cqzhao21@m.fudan.edu.cn}
\author[mymainaddress]{Zhao Yao}
\ead{20110720107@fudan.edu.cn}
\author[mymainaddress]{Zhaoyu Hu}
\ead{19210720196@fudan.edu.cn}
\author[mymainaddress]{Yuanxin Xie}
\ead{21210720255@m.fudan.edu.cn}
\author[myfourthaddress]{Yafang Zhang}
\ead{zhangyf2@sysucc.org.cn}
\author[mymainaddress]{Yuanyuan Wang}
\ead{yywang@fudan.edu.cn}
\author[mythirdaddress]{Shuo Li}
\ead{slishuo@gmail.com}
\author[myfourthaddress]{Jianhua Zhou}
\ead{zhoujh@sysucc.org.cn}
\author[myfifthaddress]{Jianqiao Zhou}
\ead{zhousu30@126.com}
\cortext[mycorrespondingauthor]{Corresponding authors: Jinhua Yu and Yin Wang}
\author[mysecondaryaddress]{Yin Wang\corref{mycorrespondingauthor}}
\ead{lpbbl@aliyun.com}
\author[mymainaddress]{Jinhua Yu\corref{mycorrespondingauthor}}
\ead{jhyu@fudan.edu.cn}

\affiliation[mymainaddress]{organization={Center for Biomedical Engineering, School of Information Science and Technology, Fudan University},
            city={Shanghai},
            postcode={200438}, 
            country={China}}
            
\affiliation[mysecondaryaddress]{organization={Department of Ultrasound, Shanghai Pulmonary Hospital, Tongji University School of Medicine},
            city={Shanghai},
            postcode={200433}, 
            country={China}}

\affiliation[mythirdaddress]{organization={Department of Biomedical Engineering and the Department of Computer and Data Science, Case Western Reserve University},
            city={Cleveland},
            postcode={OH 44106}, 
            country={USA}}
            
\affiliation[myfourthaddress]{organization={Department of Ultrasound, Sun Yat-Sen University Cancer Center, State Key Laboratory of Oncology in South China, Collaborative Innovation Center for Cancer Medicine},
            city={Guangzhou},
            postcode={510060}, 
            country={China}}
            
\affiliation[myfifthaddress]{organization={Department of Ultrasound, Ruijin Hospital, Shanghai Jiaotong University School of Medicine},
            city={Shanghai},
            postcode={200025}, 
            country={China}}  

\begin{abstract}
In the intelligent diagnosis of bimodal (gray-scale and contrast-enhanced) ultrasound videos, medical domain knowledge such as the way sonographers browse videos, the particular areas they emphasize, and the features they pay special attention to, plays a decisive role in facilitating precise diagnosis. Embedding medical knowledge into the deep learning network can not only enhance performance but also boost clinical confidence and reliability of the network. However, it is an intractable challenge to automatically focus on these person- and disease-specific features in videos and to enable networks to encode bimodal information comprehensively and efficiently. This paper proposes a novel Tri-Attention Selective Learning Network (TASL-Net) to tackle this challenge and automatically embed three types of diagnostic attention of sonographers into a mutual transformer framework for intelligent diagnosis of bimodal ultrasound videos. Firstly, a time-intensity-curve-based video selector is designed to mimic the temporal attention of sonographers, thus removing a large amount of redundant information while improving computational efficiency of TASL-Net. Then, to introduce the spatial attention of the sonographers for contrast-enhanced video analysis, we propose the earliest-enhanced position detector based on structural similarity variation, on which the TASL-Net is made to focus on the differences of perfusion variation inside and outside the lesion. Finally, by proposing a mutual encoding strategy that combines convolution and transformer, TASL-Net possesses bimodal attention to structure features on gray-scale videos and to perfusion variations on contrast-enhanced videos. These modules work collaboratively and contribute to superior performance. We conduct a detailed experimental validation of TASL-Net's performance on three datasets, including lung, breast, and liver, with a total of 791 cases. A comprehensive ablation experiment and comparison with five state-of-the-art methods resulted in diagnostic AUCs of 0.86, 0.86, and 0.97 on the three datasets, an average diagnostic accuracy improvement of $6.43\%$ over the next best method. The results show that combining medical domain knowledge with mutual spatiotemporal feature encoding can effectively improve bimodal ultrasound video diagnostic performance.
\end{abstract}




\end{frontmatter}


\section{Introduction}
Intelligent diagnosis of bimodal (gray-scale and contrast-enhanced) ultrasound (US) video relies heavily on the excellent spatiotemporal encoding performance of deep learning networks. Bimodal US video is an indispensable tool for screening and diagnosing superficial organ diseases such as lung, breast, liver, thyroid, and so on \citep{sperandeo2014transthoracic,soldati2019role,kim2017contrast, peng2020ultrasound, zhou2022contrast, sood2019ultrasound, qian2021prospective, niu2022value, alexander2020thyroid, trimboli2020performance}. As a noninvasive, radiation-free, and real-time imaging technique, bimodal US video provides a wealth of spatial and temporal information that is crucial for accurate diagnosis. Gray-scale US (GSUS) videos contain spatial information such as position, shape, texture, structure, and boundaries. It is particularly useful for identifying structural abnormalities and tissue features. However, GSUS videos exhibit high inter-frame similarity, which limits their diagnostic utility primarily to morphological information \citep{sperandeo2014transthoracic, bi2021us}. On the other hand, contrast-enhanced US (CEUS) videos provide irreplaceable functional information, such as perfusion patterns, intensity differences, perfusion times, and necrotic areas \citep{sartori2013contrast, sidhu2018efsumb}. This is crucial for analyzing blood supply differences between tissues. These complementary information improve diagnostic accuracy by comprehensively evaluating tumors \citep{du2012differentiating, yang2015effects}. To integrate these comprehensive information for precise diagnosis, deep learning networks must be able to model long-term temporal information and encode detailed spatial features simultaneously. Furthermore, bimodal US videos contain redundant information of low diagnostic value. Automated analysis and extraction of video clips with high clinical relevance are significant in improving diagnostic performance. Therefore, intelligent diagnosis of bimodal US video relies on spatiotemporal feature encoding, high-value information extraction, and high-efficient calculation. 

Despite many studies and significant efforts in deep learning-based analysis for US videos, few have focused specifically on bimodal US videos. This is partly due to the challenge of obtaining satisfactory deep learning networks with small bimodal US video datasets. Furthermore, processing bimodal videos requires more expensive computational and memory resources compared to images and unimodal videos. This limits the video input length and negatively impacts prediction accuracy. Most of existing deep learning studies for bimodal US videos typically limit the diagnosis to black-box feature extraction for given datasets, while experienced sonographers can make fairly accurate diagnoses based on their medical knowledge. Therefore, medical domain knowledge plays a decisive role in promoting reasonably accurate diagnosis for deep learning networks. The medical domain knowledge mainly includes the way sonographers browse US videos, the particular areas they emphasize, and the features they pay special attention to. It emphasizes high diagnostic value information that cannot be extracted through feature encoding alone.

\begin{figure}[!htb]
\centering
\centerline{\includegraphics[width=0.8\linewidth]{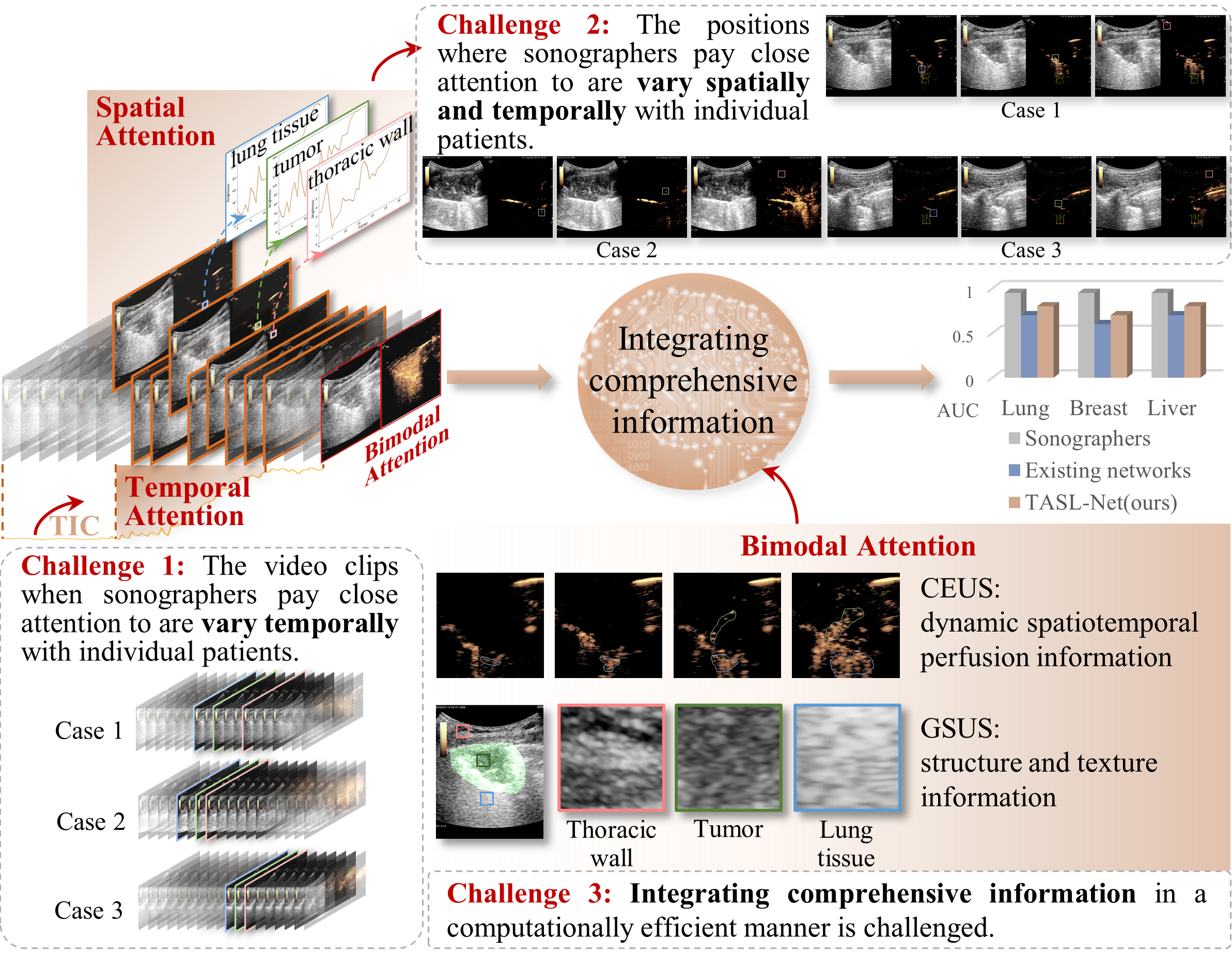}} 
\caption{Tri-attention of sonographers is crucial for accurate classification in bimodal US video. Temporal: focusing on identifying moments with rapid perfusion variations. Spatial: emphasizing perfusion differences inside and outside the tumor. Bimodal: taking into account texture and structure in GSUS, as well as long-term and dynamic perfusion variations in CEUS. Three intractable challenges, including the temporal variety of key video clips, spatiotemporal variety of key positions, and integration of comprehensive information from GSUS and CEUS videos have been overcome in this paper.}
\label{fig1}
\end{figure}
In the intelligent diagnosis of bimodal US video, the sonographers' triple diagnostic attention (Fig. 1) as the essential medical domain knowledge is as crucial as the excellent encoding performance of the deep learning network. Temporal: Sonographers typically focus their temporal attention on video clips with rapid perfusion variations. Recent medical studies \citep{jung2021quantification,schwarz2021quantitative} have shown that the time-intensity curve (TIC) is a typical factor for analyzing perfusion variations. Spatial: Sonographers highlight perfusion variations within tissue and perfusion differences between tissues \citep{caremani2008contrast, sartori2013contrast, sidhu2018efsumb, jacobsen2022clinical}. Bi et al. \citep{bi2021us} point out that \textit{the TIC of the earliest-enhanced position in each tissue is a crucial parameter for analyzing perfusion differences between tissues}. Bimodal: Sonographers selectively emphasize structure features in the GSUS videos and dynamic perfusion spatio-temporal information in the CEUS videos. For instance, in diagnosing lung tumors, sonographers scrutinize rapidly perfusion-enhanced video clips, manually select the earliest-enhanced positions on different tissues, and use TIC software to generate corresponding TICs for analyzing perfusion variations and differences. This process increases their workload and may lead to inter-observer variability. Therefore, there is an urgent need to build the bridge between clinical expertise and computer-aided diagnosis. Incorporating sonographers' diagnostic attention can help deep learning networks mimic diagnostic patterns of sonographers and focus on the moments, areas, and features they pay attention to, thereby achieving more accurate intelligent diagnosis \citep{chen2021domain}. Thus, to achieve an accurate diagnosis, it is equally essential to incorporate diagnostic attention and improve the spatiotemporal information encoding performance of deep learning networks. 

However, to our best knowledge, creating a fully automated, domain knowledge-powered deep learning network for the bimodal US video remains unaddressed. This is primarily due to the following intractable challenges. This is primarily due to the following intractable challenges (Fig. 1):
\begin{enumerate}[a.]
    \item Sonographers typically carefully scrutinize the video to identify the rapidly enhanced moments most relevant to the diagnosis. Automatically identifying these video clips is difficult because they temporally vary with individual patients.
    \item Sonographers pay the most attention to the earliest-enhanced positions to analyze perfusion differences inside and outside the lesion. These positions vary spatially and temporally between patients and, coupled with the complex patterns of perfusion variation, pose an even greater obstacle.
    \item Sonographers are concerned with tissue texture and structure in the GSUS video and dynamic perfusion variations in the CEUS video. Efficiently integrating comprehensive information from bimodal US videos without significantly adding computation complexity is also a great challenge.
\end{enumerate}

In this paper, we overcome these challenges and make the following contributions: 
\begin{itemize}
    \item We propose TASL-Net, the first fully automated domain knowledge-powered network to diagnose multiple cancers. For the first time, sonographers' tri-attention (temporal, spatial, and bimodal diagnostic attention) is simultaneously integrated into a deep learning network for accurate intelligent diagnosis.
    \item Temporal: We develop a new adaptive inflection-point detection algorithm to automatically select video clips characterized by dynamic perfusion enhancement. This method effectively addresses patient-specific variations and improves computational efficiency.
    \item Spatial: We design a new optimal feature representation method to highlight TIC features at the earliest-enhanced positions emphasized by sonographers. This method guides the TASL-Net to dynamically analyze patient-specific perfusion variations and differences.
    \item Bimodal: We propose a new mutual encoding strategy to capture texture and structure in GSUS and dynamic perfusion information in CEUS. This strategy facilitates feature extraction performance by effectively addressing both detailed spatial information within frames and long-term temporal dependence between sequential frames.
\end{itemize}

To evaluate the performance of TASL-Net, we have conducted detailed and sufficient experiments on lung, breast, and liver datasets, totaling 791 cases. The results have demonstrated that TASL-Net has achieved satisfying performance on these datasets, without requiring manual intervention in the diagnostic process. Our approach offers a promising solution for the intelligent diagnosis of multiple cancers using bimodal US video, alleviating burdens on sonographers and avoiding intra-observer variability.

\section{Related Works}
\subsection{Intelligent Diagnosis Based on US Videos}
Although there is a limited number of deep learning works on the analysis of bimodal US videos, many studies have significantly contributed to intelligent diagnosis using US videos. Our work is inspired by the substantial progress of these deep learning studies. Ebadi et al. \citep{ebadi2021automated} introduced a deep learning technique for classifying lung US video scans acquired at point-of-care without requiring any further processing or operator intervention. C{\u{a}}leanu et al. \citep{cualeanu2021deep} examine the use of CEUS video with deep neural networks for automated liver tumor diagnosis. Schmiedt et al. \citep{schmiedt2022preliminary} propose a deep learning approach for the fully automated intelligent diagnosis of CEUS video, allowing for the identification of specific focal liver tumors. For bimodal US video, Yang et al. \citep{yang2020temporal} design a temporal sequence dual-branch network for classifying bimodal video of breast cancer, while Gong et al. \citep{gong2022bus} propose a bimodal US network called BUS-net for breast cancer diagnosis. These works have made significant contributions to the development of the intelligent diagnosis of US videos. 

\subsection{Domain Knowledge Powered US Data Analysis}
According to recent research in deep learning, embedding medical domain knowledge is beneficial in accurately diagnosing various diseases. Xie et al. \citep{xie2021survey} provide a comprehensive review of the studies embedding medical domain knowledge into networks for various medical tasks. They analyze the current research status, provide a basis for new research, and help researchers to understand the embedding of medical domain knowledge better. In the field of US image analysis, Yang et al. \citep{yang2021integrate} incorporate a $1 \times 9$ medical feature vector, including dimension, growth, aspect ratio, and other relevant features, to train a multi-task cascade network for the benign versus malignant classification of thyroid nodules. Frank et al. \citep{frank2021integrating} consider features of anatomical phenomena, such as the pleural line and presence of consolidations, as well as sonographic artifacts, such as A- and B-lines, as feature maps, and integrate them into deep neural networks for lung US classification to predict COVID-19 severity. They demonstrate that the spatial attention of sonographers can effectively improve the network's performance. In US video analysis, Chen et al. \citep{chen2021domain} incorporate the specific time slots when radiologists browsed CEUS video and the different boundaries on CEUS and GSUS frames into the network for diagnosing breast cancer based on bimodal (GSUS and CEUS) videos. Their well-executed experiments and excellent results demonstrate that domain knowledge can effectively enhance the video understanding performance of the network. Furthermore, they also provide a new idea and method for intelligent breast cancer diagnosis. These domain knowledge-powered analysis works have inspired our research.

\subsection{Transformer Cooperates with Convolution for Video Understanding}
The Transformer is a powerful framework for video analysis due to its ability to capture global dependence over long-term features \citep{selva2022video}. Bertasius et al. \citep{bertasius2021space} propose a convolution-free model called TimeSformer for video classification, comparing the performance of five space-time self-attention schemes. Arnab et al. \citep{arnab2021vivit} also present a pure-transformer-based model called ViViT for video classification. They design several efficient variants of their model to handle long sequences of video and achieve SOTA results on multiple video classification benchmarks. However, these models suffer from expensive computation. Liu et al. \citep{liu2022video} advocate for the inductive bias in transformers and propose a spatiotemporal adapted network called video swin transformer (VST). The VST leads to a better trade-off between speed and accuracy for video analysis than previous approaches. However, pure transformer networks cannot encode high-level local information, which is precisely necessary for extracting features of perfusion variations in bimodal US video \citep{dai2021coatnet}.

The combination of convolution and transformer has proven to be highly effective in fusing global, local, temporal, and spatial information \citep{wu2021cvt, peng2021local}. Many studies have demonstrated performance improvements by combining these two frameworks. For example, Girdhar et al. \citep{girdhar2019video} developed an action transformer network that learns spatiotemporal context in video clips for localizing and classifying human actions. Liu et al. \citep{liu2020convtransformer} introduced ConvTransformer, a novel end-to-end network for learning and synthesizing video frame sequences. Feng et al. \citep{feng2021convolutional} proposed convolutional transformer-based dual discriminator generative adversarial networks (CT-D2GAN) for unsupervised video anomaly detection. Peng et al. \citep{peng2021local} propose a hybrid network called Conformer to take advantage of convolutional operations and self-attention mechanisms for enhanced representation learning. Inspired, we design a mutual encoding framework by combining convolution and transformer. Leveraging the strengths of both approaches, our framework efficiently encodes the spatial expression within each frame while capturing the dynamic temporal variability between sequential frames.

\section{TASL-Net}
\subsection{Method Overview}
\begin{figure}[!htb]
\centering
\centerline{\includegraphics[width=0.8\linewidth]{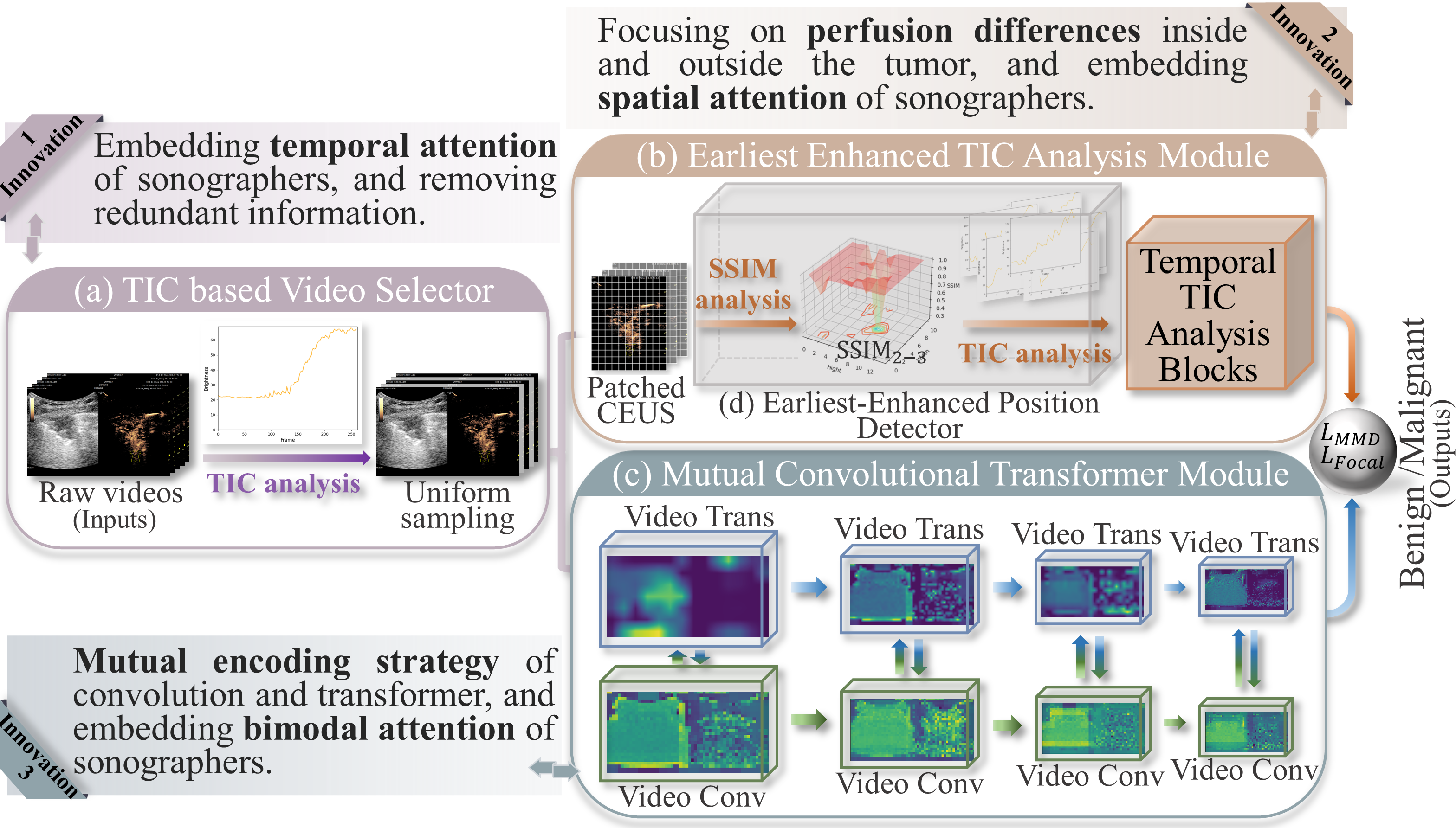}} 
\caption{The TASL-Net is composed of three key components: (a) TIC-based Video Selector; (b) ETIC-Earliest Enhanced TIC Analysis Module; (c) MCT-Mutual Convolutional Transformer Module. They collectively address a previously unsolved challenge: analyzing the unique perfusion characteristics specific to each patient for accurate intelligent diagnosis.}
\label{fig2}
\end{figure}
The TASL-Net is an automated intelligent diagnosis network that embeds the tri-attention of sonographers and integrates comprehensive information from bimodal US videos. It is composed of three key components, all collectively contributing to \textit{Tri-Attention Selective Learning}. Firstly, the TIC-based video selector (Fig. 2 (a)) samples the key video clips of rapid perfusion-enhancement, as accurately as a sonographer would. Then, the earliest-enhanced TIC analysis module (ETIC, Fig. 2 (b)) automatically incorporates perfusion differences inside and outside the tumor by paying close attention to the features in sonographers emphasized positions (i.e., the earliest-enhanced positions). Meanwhile, the mutual convolutional transformer module (CMT, Fig. 2 (c)) extracts the texture and structure features and dynamic perfusion information in the video by mutual encoding strategy between convolution and transformer. The ETIC and CMT modules are jointly optimized to match their feature distributions. These three components collectively address a previously unsolved challenge: analyzing the unique perfusion characteristics specific to each patient for accurate intelligent diagnosis in bimodal ultrasound videos. \textit{This paper deals with the diagnosis of lung, breast, and liver cancers. To ensure clarity and provide detailed explanations, we use lung tumors as an example in this section.}

\subsection{TIC-based Video Selector}
\begin{figure}[!htb]
\centering
\centerline{\includegraphics[width=0.7\linewidth]{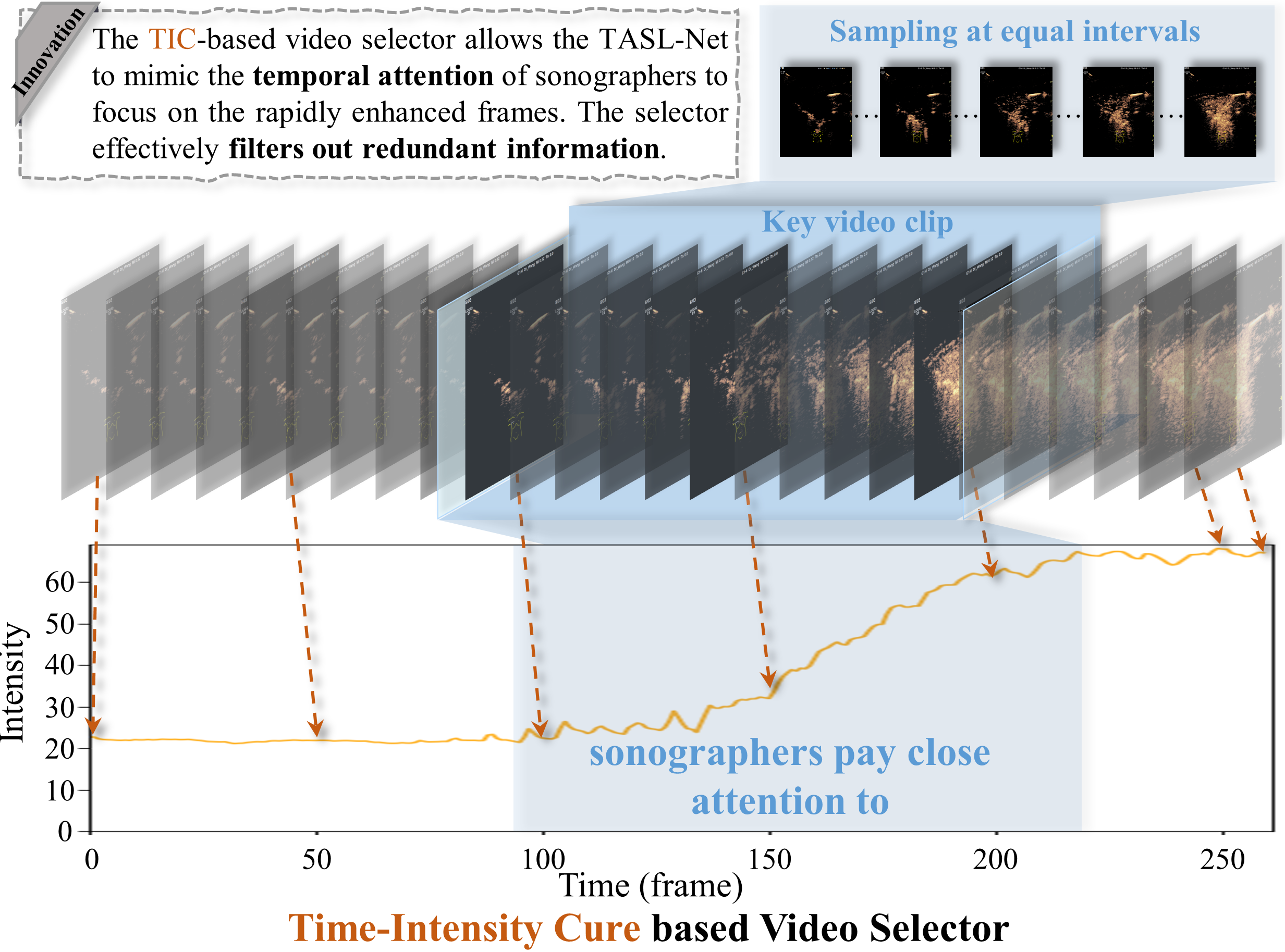}}
\caption{The TIC-based video selector automatically extracts video clips that sonographers pay close attention to, surpassing the limitations of conventional sampling methods. It improves computational efficiency by filtering out redundant information.}
\label{fig3}
\end{figure}

The TIC-based video selector (Fig. 3) embeds the temporal attention of sonographers into the TASL-Net to detect the perfusion-enhanced moments in the video. The selector analyzes the TIC of the CEUS video and adaptively selects the rapidly rising portion. By eliminating redundant information, the TIC-based selector improves the efficiency of video interpretation and reduces the computational load of data processing. Additionally, it is flexible enough to handle input videos of varying lengths, adapting to patient-specific variations, and tailoring the selection process for each individual.

\begin{breakablealgorithm}
\small
  \caption{TIC based selection strategy}
  \label{alg1}
  \begin{algorithmic}[1]
    \Require
      The original bimodal videos $\mathcal{V}_{BUS}$; Video length $F_0$; Window length $\varepsilon$; Poly order $\eta$; Gradient threshold $\delta$
    \Ensure
      The sampled bimodal videos $\mathcal{V}$
    \State Calculate TIC of the $\mathcal{V}_{BUS}$;
    \State $\mathcal{C} = S-G(\mathcal{C}, \varepsilon, \eta)$;
    \For f
    \If {$ \frac{\mathrm{d} c^f}{\mathrm{d} f} > \delta$}
        \State $t_{TTS}$ = $f$
        \EndIf
        \State max = $c^f$
        \If {$c^{f+1} > max$}
        \State max = $c^{f+1}$
        \State $t_{TTP}$ = $f+1$
        \EndIf
    \EndFor
    \State $quo, rem = (t_{TTP}-t_{TTS}) \div F$
    \State $\Delta = quo+1$
    \State $t_{TTP}=F-rem+t_{TTP}$
    \State $\mathcal{V} = \{v_{bus}^f|f = t_{TTS},t_{TTS}+\Delta, t_{TTS}+2\Delta,..., t_{TTP}\}$
  \end{algorithmic}
\end{breakablealgorithm}

Given a raw bimodal US video $\mathcal{V}_{BUS}=\{v_{bus}^f|f = 1,2,...,F_0\}$, conventional preprocessing methods that uniformly sample the entire video often fail to filter out redundant information. In contrast, the TIC-based video selector employs an adaptive algorithm (Algorithm 1) to select key video clips and remove redundant information. Firstly, the selector analyzes the average intensity $c^f$ per frame of the entire video to generate a TIC $\mathcal{C}=\{c^f|f = 1,2,...,F_0\}$ that shows how the intensity changes over time. Then, the TIC is smoothed by the Savitzky-Golay filter ($S-G()$) \citep{schafer2011savitzky} to identify the start-to-enhanced point $t_{TTS}$ and peak point $t_{TTP}$ on the curve, and the key video clips are extracted based on these points. The window length $\varepsilon$ refers to the number of intensity points in the window used for polynomial fitting. The polynomial order $\eta$ determines the degree of the polynomial used to fit the data within the window. It fits complex, nonlinear data within the window, enhancing the detection of subtle changes in intensity. According to \citep{xu2020savitzky}, we set $\varepsilon = 31$ and $\eta = 2$ to eliminate noise, identify significant trends in the TIC, and preserve more detail. The gradient threshold $\delta$ sets the sensitivity for detecting changes in the TIC. We set $\delta = 0.2$ to minimize the impact of minor fluctuations, focusing on significant changes. Finally, the selector samples $F=32$ frames from the key video clips at equal intervals to compromise lower computation and memory requirements, which are used as input for the ETIC and CMT modules.

\subsection{Earliest-Enhanced TIC Analysis}
The earliest-enhanced TIC analysis module (ETIC) automatically identifies the earliest-enhanced positions where sonographers focus their spatial attention to emphasize the perfusion variations within tissue and perfusion differences between tissues. As shown in Fig. 2 (b), the ETIC consists of two components: the earliest-enhanced position detector (Fig. 2 (d)) and the temporal TIC analysis blocks (Fig. 5 (b)).

\subsubsection{SSIM-based Earliest Enhanced Position Localization}
\begin{figure}[!htb]
\centering
\centerline{\includegraphics[width=0.8\linewidth]{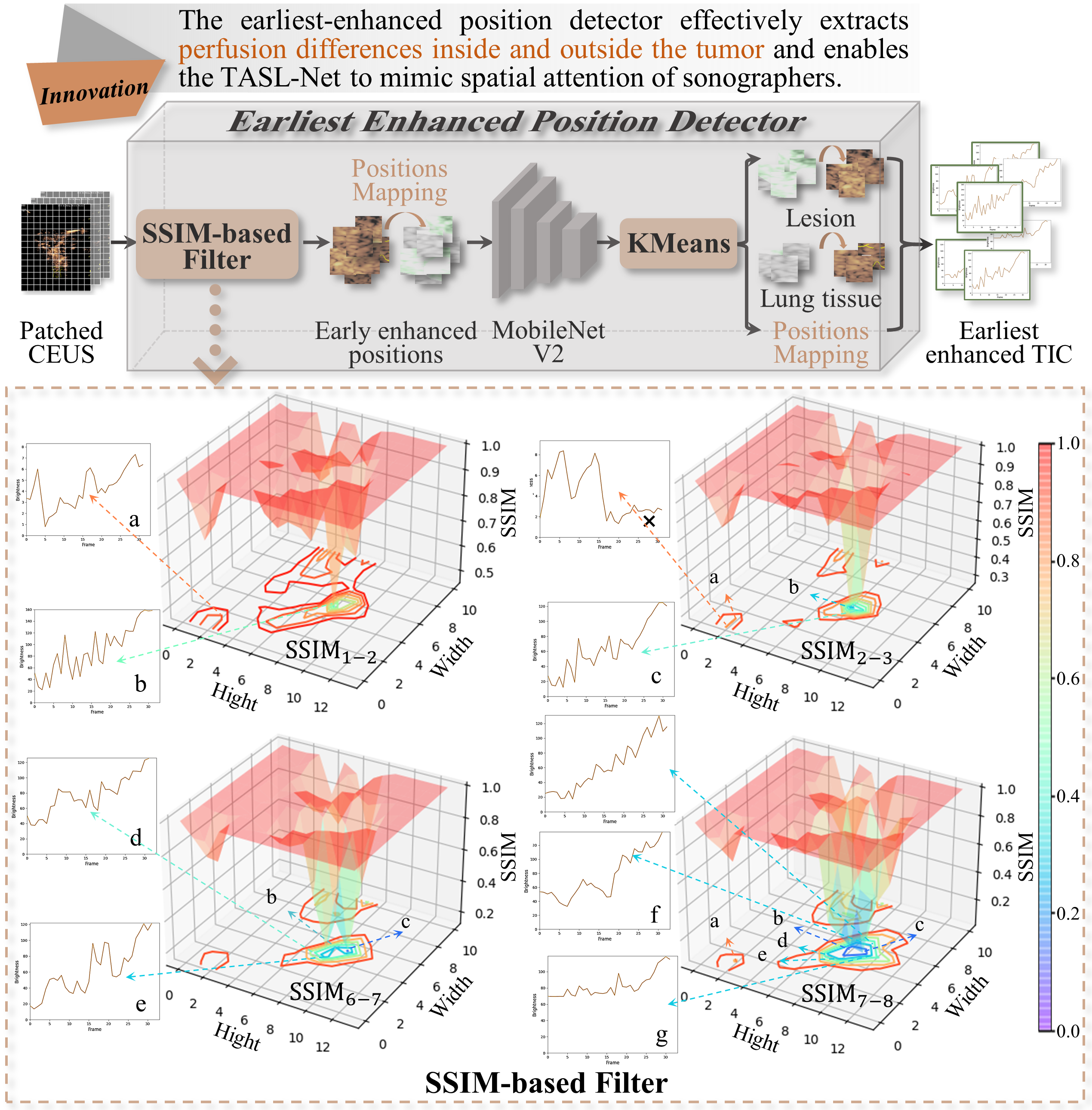}} 
\caption{The earliest-enhanced position detector has accurately analyzed the structural similarity (SSIM) changing and automatically selected the key position of great clinical attention. The ``$SSIM_{*-**}$'' denote the SSIM between the $*^{th}$ and $**^{th}$ frame. The markers "a, b, c, ..." denote patches with low SSIM values, representing early-enhanced positions in each tissue. These markers correspond to those in Fig. 6.}
\label{fig4}
\end{figure}

The earliest-enhanced position detector (Fig. 4) automatically identifies the key positions in different tissues based on the structural similarity (SSIM) \citep{wang2004image} changing. The input to the detector are $H \times 2W \times F$ sampled bimodal US videos $\mathcal{V}=\{v_{bus}^f|f=1,2,...,F\}$ consisting of GSUS videos and CEUS videos $\mathcal{V}_{CE}=\{v_{ce}^f|f=1,2,...,F\}$, where $H=896$ and $W=704$ are height and width of GSUS and CEUS videos. Firstly, the detector divides the $\mathcal{V}_{CE}$ into a patch set ${\mathcal{P}}=\{p_{i,j}^f|i=1,2,...,\frac{H}{64}, j=1,2,...,\frac{W}{64}, f=1,2,...,F\}$, where the $p_{i,j}^f$ represents the patch at the $i^{th}$ row and the $j^{th}$ column on the $f^{th}$ frame. The patch size is $64\times64$. Then, the detector calculates the patch-level SSIM to reflect the perfusion enhancement. The SSIM value set between two temporal-adjacent frames can be derived as follows:
\begin{equation}
\begin{array}{l}
SSIM(v_{CE}^f, v_{CE}^{f+1}) = \\
\bigcup_{i=1, j=1,1}^{i=\frac{H}{64}, j=\frac{W}{64}}\frac{(2\mu_{p_{i,j}^f}\mu_{p_{i,j}^{f+1}}+\epsilon_1)(2\sigma_{p_{i,j}^fp_{i,j}^{f+1}}+\epsilon_2)}{(\mu_{p_{i,j}^f}^2+ \mu_{p_{i,j}^{f+1}}^2+\epsilon_1)(\sigma_{p_{i,j}^f}^2+\sigma_{p_{i,j}^{f+1}}^2+\epsilon_2)},
\end{array}
\end{equation}
where $\epsilon_1=(0.01R)^2$, $\epsilon_2=(0.03R)^2$, and $R=255$ is the range of pixel values. For the temporal-adjacent patches with same spatial position, $\mu_{p_{i,j}^f}$ and $\mu_{p_{i,j}^{f+1}}$ denote their expected value, $\sigma_{p_{i,j}^f}$ and $\sigma_{p_{i,j}^{f+1}}$ denote their variance, and $\sigma_{p_{i,j}^fp_{i,j}^{f+1}}$ denote their covariance.
\begin{breakablealgorithm}
\small
  \caption{Earliest-enhanced TICs calculation}
  \label{alg2}
  \begin{algorithmic}[1]
    \Require
      The sampled bimodal videos $\mathcal{V}$; Hyper-parameter $\epsilon_1$ and $\epsilon_2$; SSIM threshold $\tau$
    \Ensure
      The earliest-enhanced TICs $\mathcal{C}$ of different tissue
    \State Partition the $\mathcal{V}$ into $64 \times 64$ and build the CEUS patch set $\mathcal{P}$;
    \For {i, j}
        \If {$SSIM(p_{i,j}^f, p_{i,j}^{f+1}) \leq \tau$}
            \If {TIC of $p_{i,j}$ is rising}
                \State Patch $p_{i,j}$ contains the early enhanced position;
            \Else
                \State The low SSIM of $p_{i,j}$ is caused by the respiratory movement;
            \EndIf
        \Else
            \State Ignore the patch $p_{i,j}$;
        \EndIf
    \EndFor
    \State KMeans(MobileNetV2(GSUS patches on early enhanced position (i, j)));
    \For {wall group, tumor group, lung tissue group}
        \State Calculate TIC of the $p_{i,j}$;
        \State Calculate $t_{TTS}$ on TIC of $p_{i,j}$;
        \State Determine two TICs with top-two enhanced $t_{TTS}$.
    \EndFor
  \end{algorithmic}
\end{breakablealgorithm}

A lower SSIM value indicates more significant changes in intensity, texture, and structure in the corresponding spatial position, that is, a greater perfusion variation. As presented in Algorithm 2, the detector adopts a low-SSIM-pass strategy to identify earliest-enhanced positions. As the thoracic wall is far less enhanced than the lung tissue and lesion, the detector divides a CEUS frame into two parts: the wall and the non-wall. For the thoracic wall, the detector selects two positions with the lowest SSIM values $\tau_{wall}=0.8$ as the earliest-enhanced positions. For the non-wall part, the detector selects all positions with an SSIM value less than $\tau_{nwall}=0.6$ as early-enhanced positions. This division ensures that the TIC analysis is performed on the appropriate regions of interest, taking into account the differential perfusion characteristics in different tissues. The SSIM changing visualization of several adjacent frames is illustrated in Fig. 4. 

To ensure a balanced number of earliest-enhanced positions and mitigate the risk of them congregating in a single anatomical structure, it is essential to classify the above early-enhanced positions into lesion or lung tissue groups. The GSUS and CEUS videos are acquired from the same anatomical structure, allowing for spatial correlation and position sharing between the two modalities. Due to the absence of structure and texture features in early-enhanced CEUS frames, the classification of these positions is carried out using GSUS frames. The detector uses cascaded MobileNetV2 \citep{sandler2018mobilenetv2} and K-means for feature encoding and binary clustering to classify the positions. 

Before being embedded into the proposed TASL-Net, MobileNetV2 undergoes pre-training. We randomly extract two $64\times64\times10$ sub-videos in the tumor and two $64\times64\times10$ sub-videos in the lung tissue from 100 cases (50 benign $+$ 50 malign) of the lung dataset. Therefore the pre-training dataset contains 4000 tumor and lung tissue patches. The pre-trained MobileNetV2 is then employed to extract spatial features from early-enhanced positions. Finally, these features are input into the K-means unit to classify the positions into either the lesion or lung tissue group. 

Once the classification of early-enhanced positions is done, the detector automatically selects the top two enhanced positions in each group as the earliest-enhanced positions. To reduce the coexistence error, the detector selects a half-sized sub-patch in the center of each earliest-enhanced position to calculate the TICs and generates the earliest-enhanced TIC sets $\mathcal{C}=\{c_n|n=1,2,...N\}=\{c_n^f|f=1,2,...,F, n=1,2,...,N\}$, where $N=6$ is the number of positions. Consequently, the detector achieves the quantification of perfusion variations and differences by extracting the earliest-enhanced TICs without the manual intervention of sonographers.

\subsubsection{Intelligent Analysis of Earliest-Enhanced TICs}
\begin{figure}[!htb]
\centering
\centerline{\includegraphics[width=1.0\linewidth]{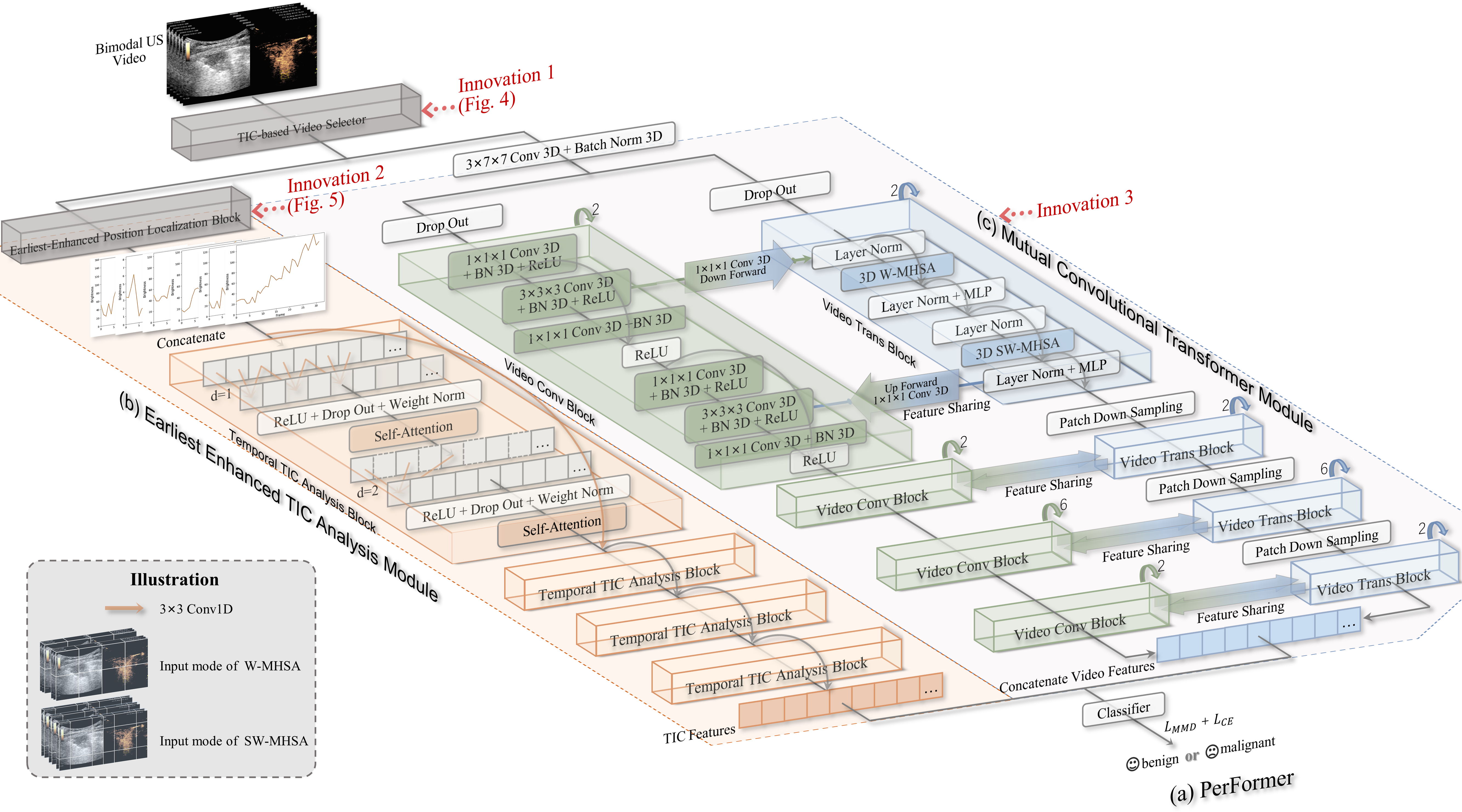}} 
\caption{With the cooperation of the three novel modules, TASL-Net takes advantage of temporal attention guidance (TIC-based Video selector), spatial attention guidance (b), mutual encoding strategy (c), and bimodal (gray-scale and contrast-enhanced) analysis for a SOTA diagnosis performance.}
\label{fig5}
\end{figure}
The cascaded temporal TIC analysis blocks (Fig. 5 (b)) capture the temporal information in the earliest-enhanced TICs. The backbone of the blocks is the temporal convolutional network (TCN, \citep{bai2018empirical}), which surpasses recurrent canonical networks such as LSTMs in encoding long-term temporal sequences while keeping the computational load low. The dilated causal convolution layers with dilation factors $d=1,2,4$ ensure a positive correlation between time and intensity. For earliest-enhanced TIC sets $\mathcal{C}$, to define the dilated convolution on an element $c_n^f$ using a filter $\phi:\{0,1,...,k-1\}\rightarrow\mathbb{R}$, it has:
\begin{equation}
\begin{array}{r}
G(c_n^f)=\sum\limits^{k-1}_{\beta=0}g(\beta)\cdot c_n(c_n^f-d\cdot \beta),
\end{array}
\end{equation}
where $k=3$ is the kernel size, the $c_n^f-d\cdot \beta$ denotes the past direction, and $\beta=0,1,2$. 

We advance the backbone by embedding hierarchical self-attention between the dilated convolution to calculate the optimal temporal representation. The convolution-encoded features $in_{TIC}$ are transformed into query $q_{TIC}$, key $k_{TIC}$ and value $v_{TIC}$ using $1\times 1$ convolution layers. The self-attention map $A_{TIC}$ is then calculated as follows:
\begin{equation}
\begin{array}{r}
A_{TIC}=softmax(q_{TIC}^T \cdot k_{TIC}).
\end{array}
\end{equation}
Finally, the optimal representation $out_{TIC}$ of earliest-enhanced TICs is obtained through:
\begin{equation}
\begin{array}{r}
out_{TIC}=concatnate(in_{TIC}, A_{TIC}^T \cdot v_{TIC}).
\end{array}
\end{equation}
The self-attention weights determine the significance of the TICs in the tumor diagnosis task. The self-attention layer guides the ETIC module to focus on the high-weighted intensity information and restrains low-value temporal expression.

\subsection{Mutual Encoding of Convolution and Transformer}
The proposed mutual convolutional transformer module (CMT, Fig. 5 (c)) enables the TASL-Net to capture the comprehensive information in GSUS and CEUS videos based on the bimodal attention of sonographers. It successfully captures detailed spatial expression within each frame and dynamic temporal variability between sequential frames, achieving a balance between feature encoding and computational efficiency.

As detailed in Fig. 5 (c) and Table 1, the CMT is composed of a stem block, a video convolution branch, and a video transformer branch. The stem block includes a $2\times4\times4$ convolution with a stride of $2\times4\times4$, followed by 3D batch normalization. The stem block is used to extract initial local features such as edges and texture information. CMT adopts four-stage feature mutual. The basic layer numbers in four stages are $(2, 2, 6, 2)$. CMT progressively feeds global context to the convolution branch and delivers detailed local information to the transformer branch.  In each layer, the residual connection is applied to address the issues of vanishing gradients and weight matrix degradation. The video convolution block captures the local spatial expression, such as structure, edge, texture, and semantic features of the lung tissue and tumor. The computational formula of each block is defined as follows:
\begin{equation}
VCB = \left\{
\begin{array}{l}
fc_1=ReLU(BN(Conv3D_1(fc_0))),\\
fc_2=ReLU(BN(Conv3D_3(fc_1))),\\
fc_3=ReLU(BN(Conv3D_1(fc_2))+fc_0),\\
fc_4=ReLU(BN(Conv3D_1(fc_3))),\\
fc_5=ReLU(BN(Conv3D_3(fc_4))),\\
fc_6=ReLU(BN(Conv3D_1(fc_5))+fc_3),\\
\end{array}
\right.
\end{equation} 
where $fc_0$ to $fc_6$ denote the convolution feature maps, $Conv3D_1()$ is a $1\times1\times1$ 3D convolution layer, $Conv3D_3()$ is a $3\times3\times3$ 3D convolution layer, $BN()$ is 3d batch normalization, and $ReLU()$ denotes a rectified linear unit as the activation function.

The video transformer block models long-term temporal information to encode the correlation between structural and intensity changes of perfusion over time.
The key components of each transformer block are the 3d window multi-head self-attention (W-MHSA) unit, along with the 3d shifted window multi-head self-attention (SW-MHSA) unit. The computational formula for each video transformer block can be defined as follows:
\begin{equation}
VTB = \left\{
\begin{array}{l}
ft_1=3dW-MHSA(LN(ft_0))+ft_0,\\
ft_2=MLP(LN(ft_1))+ft_1,\\
ft_3=3dSW-MHSA(LN(ft_2))+ft_2,\\
ft_4=MLP(LN(ft_3))+ft_3,
\end{array}
\right.
\end{equation}
where the $ft_0$ to $ft_4$ represent the transformer feature tokens, $MLP()$ is a multilayer perceptron unit, and $LN()$ denotes layer normalization. The patch down-sampling block performs spatial downsampling between the two stages.

\begin{table}[!htp]
\Huge
\centering
\caption{Architecture of TASL-Net. The ``$MHSA-*$'', ``$W-MHSA-*$'', and ``$SW-MHSA-*$'' denote the multi-head self-attention with * heads. In the feature sharing column, the arrows represent the flow of features. In output column, the Numbers in ``()'' represent the channel, height, width and dimension of features.}\label{tab1}
\renewcommand\arraystretch{1.4}
\smallskip\resizebox{0.95\linewidth}{85mm}{\begin{tabular}{c|c|c|c|c|c|c}
\hline
\multirow{2}{*}{stage} & \multirow{2}{*}{TIC Branch} & \multicolumn{3}{c|}{$V_{BUS}$ Branch [$s1\times2, s2\times2, s3\times6, s4\times2$]} & \multicolumn{2}{c}{output} \\
\cline{3-7}
\multirow{1}{*}{} & \multirow{1}{*}{} & Conv Branch & Feature Sharing & Trans Branch & TIC & $V_{BUS}$ \\
\hline
pre-processing & \multicolumn{4}{c|}{Algorithm 1} & \multicolumn{2}{c}{(32, 224, 448, 3)} \\
\hline
s0 & Algorithm 2 & \multicolumn{3}{c|}{$2\times4\times4$, 96, stride=(2,4,4)} & (6, 32) & (8, 56, 112, 96) \\
\hline
\multirow{7}{*}{$s1$} &  & $1\times1\times1, 64$ & \multirow{3}{*}{$1\times1\times1, 96 \rightarrow$} &  & \multirow{6}{*}{(32, 25)} & \multirow{6}{*}{(8, 56, 112, 96)} \\
\multirow{6}{*}{} & conv1d-2, 25 & $3\times3\times3, 64$ & \multirow{2}{*}{} &  & \multirow{5}{*}{} & \multirow{5}{*}{} \\
\multirow{5}{*}{} & MHSA-3 & $1\times1\times1, 128$ & \multirow{1}{*}{} & W-MHSA-3 & \multirow{4}{*}{} & \multirow{4}{*}{}  \\
\cdashline{3-3}
\multirow{4}{*}{} & conv1d-2, 25 & $1\times1\times1, 64$ & \multirow{3}{*}{$\leftarrow 1\times1\times1, 64$} & SW-MHSA-3 & \multirow{3}{*}{} & \multirow{3}{*}{} \\
\multirow{3}{*}{} & MHSA-3 & $3\times3\times3, 64$ & \multirow{2}{*}{} &  & \multirow{2}{*}{} & \multirow{2}{*}{} \\
\multirow{2}{*}{} &  & $1\times1\times1, 128$ & \multirow{1}{*}{} &  & \multirow{1}{*}{} & \multirow{1}{*}{} \\
\cline{2-7}
\multirow{1}{*}{} & \multicolumn{3}{c|}{-} & patch down-sampling & - & (8, 28, 56, 192) \\
\hline
\multirow{7}{*}{$s2$} &  & $1\times1\times1, 128$ & \multirow{3}{*}{$1\times1\times1, 192 \rightarrow$} &  & \multirow{6}{*}{(32, 25)} & \multirow{6}{*}{(8, 28, 56, 192)} \\
\multirow{5}{*}{} & conv1d-2, 25 & $3\times3\times3, 128$ & \multirow{2}{*}{} &  & \multirow{6}{*}{} & \multirow{5}{*}{} \\
\multirow{5}{*}{} & MHSA-3 & $1\times1\times1, 256$ & \multirow{1}{*}{} & W-MHSA-6 & \multirow{4}{*}{} & \multirow{4}{*}{}  \\
\cdashline{3-3}
\multirow{4}{*}{} & conv1d-2, 25 & $1\times1\times1, 128$ & \multirow{3}{*}{$\leftarrow 1\times1\times1, 128$} & W-SMHSA-6 & \multirow{3}{*}{} & \multirow{3}{*}{} \\
\multirow{3}{*}{} & MHSA-3 & $3\times3\times3, 128$ &  & \multirow{2}{*}{} & \multirow{2}{*}{} & \multirow{2}{*}{} \\
\multirow{2}{*}{} &  & $1\times1\times1, 256$ & \multirow{1}{*}{} &  & \multirow{1}{*}{} & \multirow{1}{*}{} \\
\cline{2-7}
\multirow{1}{*}{} & \multicolumn{3}{c|}{-} & patch down-sampling & - & (8, 14, 28, 384) \\
\hline
\multirow{7}{*}{$s3$} &  & $1\times1\times1, 256$ & \multirow{3}{*}{$1\times1\times1, 384 \rightarrow$} &  & \multirow{6}{*}{(32, 25)} & \multirow{6}{*}{(8, 14, 28, 384)} \\
\multirow{6}{*}{} & conv1d-2, 25 & $3\times3\times3, 256$ & \multirow{2}{*}{} &  & \multirow{5}{*}{} & \multirow{5}{*}{} \\
\multirow{5}{*}{} & MHSA-3 & $1\times1\times1, 512$ & \multirow{1}{*}{} & W-MHSA-12 & \multirow{4}{*}{} & \multirow{4}{*}{}  \\
\cdashline{3-3}
\multirow{4}{*}{} & conv1d-2, 25 & $1\times1\times1, 256$ & \multirow{3}{*}{$\leftarrow 1\times1\times1, 256$} & SW-MHSA-12 & \multirow{3}{*}{} & \multirow{3}{*}{} \\
\multirow{3}{*}{} & MHSA-3 & $3\times3\times3, 256$ & \multirow{2}{*}{} &  & \multirow{2}{*}{} & \multirow{2}{*}{} \\
\multirow{2}{*}{} &  & $1\times1\times1, 512$ & \multirow{1}{*}{} &  & \multirow{1}{*}{} & \multirow{1}{*}{} \\
\cline{2-7}
\multirow{1}{*}{} & \multicolumn{3}{c|}{-} & patch down-sampling & - & (8, 7, 14, 768) \\
\hline
\multirow{6}{*}{$s4$} &  & $1\times1\times1, 256$ & \multirow{3}{*}{$1\times1\times1, 768 \rightarrow$} &  & \multirow{6}{*}{(32, 25)} & \multirow{6}{*}{(8, 7, 14, 768)} \\
\multirow{5}{*}{} & conv1d-2, 25 & $3\times3\times3, 256$ & \multirow{2}{*}{} &  & \multirow{5}{*}{} & \multirow{5}{*}{} \\
\multirow{4}{*}{} & MHSA-3 & $1\times1\times1, 512$ & \multirow{1}{*}{} & W-MHSA-24 & \multirow{4}{*}{} & \multirow{4}{*}{}  \\
\cdashline{3-3}
\multirow{3}{*}{} & conv1d-2, 25 & $1\times1\times1, 256$ & \multirow{3}{*}{$\leftarrow 1\times1\times1, 256$} & SW-MHSA-24 & \multirow{3}{*}{} & \multirow{3}{*}{} \\
\multirow{2}{*}{} & MHSA-3 & $3\times3\times3, 256$ & \multirow{2}{*}{} &  & \multirow{2}{*}{} & \multirow{2}{*}{} \\
\multirow{1}{*}{} &  & $1\times1\times1, 512$ & \multirow{1}{*}{} &  & \multirow{1}{*}{} & \multirow{1}{*}{} \\
\hline
\multirow{2}{*}{classifier} & \multicolumn{4}{c|}{average pooling} & \multicolumn{2}{c}{\multirow{2}{*}{$1\times1, 1$}} \\
 & \multicolumn{4}{c|}{fully connected} & \multicolumn{2}{c}{} \\
\hline
\end{tabular}}
\end{table}

The semantic gap between convolution and transformer features is considerable, with the former primarily expressing local spatial information and the latter long-term temporal information. To address this, we design the feature sharing unit to bridge the gap and improve CMT’s ability to learn the spatial correspondence between GSUS and CEUS videos. For simplicity, this paragraph mainly illustrates the cases from convolution to transformer blocks, which can easily be generalized to the reverse cases. To accomplish this, the shared convolution features can be defined as:
\begin{equation}
\begin{array}{l}
f_{c\rightarrow t}=Conv3D_1(Reshape(fc_2)),
\end{array}
\end{equation}
where the $Reshape()$ denotes the dimensions match of two features and $fc_2$ denotes the output of the second convolution layer in a video convolution block. This unit builds the feature fusion bridges between convolution and transformer. This is particularly useful for distinguishing perfusion variations on different anatomical structures in CEUS videos.

The mutual encoding strategy enhances the encoding of detailed spatial features within frames and captures dynamic temporal variability across successive frames. Its progressive feature-sharing mechanism facilitates multi-scale information interaction between GSUS and CEUS modalities, optimizing both computational and memory efficiency.

\subsection{TASL-Net Optimization}
\begin{breakablealgorithm}
\small
  \caption{TASL-Net Optimization}
  \label{alg3}
  \begin{algorithmic}[1]
    \Require
      The original bimodal US videos $\mathcal{V}_{BUS}$; Video length $F_0$; Hyper-parameter $\epsilon_1$ and $\epsilon_2$; Window length $\varepsilon$; Poly order $\eta$; Gradient threshold $\delta$; SSIM threshold $\beta$; The number of Temporal TIC Analysis Block $K$; The cascade times M in CMT module; The label of the types of disease $G$; The loss balanced weights $\lambda$; Train epoch $Ep$; Batch size $Bs$; Learning rates $Lr$ 
    \Ensure
      Learned parameters {$\theta_{ETIC}$, $\theta_{CMT}$}; The diagnostic result
    \State \textbf{Algorithm 1};
    \State Initialize the parameters {$\theta_{ETIC}$, $\theta_{CMT}$};
    \For {epoch}          
        \State fed $v^f, c_n, G$ $\leftarrow v^f$ and $c_n$ represent the $\mathcal{V}$ and $\mathcal{C}$ of each batch;
        \State /*\textbf{Forward} propagation of \textbf{mutual} branch 1: \textbf{ETIC}*/
        \State \textbf{Algorithm 2};
        \For {K}
            \State $Conv1D = G(c_n)$;
            \State $A_{TIC} = Self-Attention(c_n)$;
        \EndFor
        \State $ETIC(c_n) = A_{TIC} + Conv1D$;
        \State /*\textbf{Forward} propagation of \textbf{mutual} branch 2: \textbf{CMT}*/
        \State $Conv3d = CNN(v^f)$;
        \For {M}
            \State $v_{CNN}^f = VCB(v^f)$;
            \State $v_{C\rightarrow T}^f = Conv3d(v_{CNN}^f)$;
            \State $v_{Tra}^f = VTB(v^f)$;
            \State $v_{T\rightarrow C}^f = Conv3d(v_{Tra}^f)$;
            \State $v_{CNN}^f = VCB(v^f + v_{T\rightarrow C}^f)$;
            \State $v_{Tra}^f = VTB(v^f + v_{C\rightarrow T}^f)$;
        \EndFor
        \State $CMT(v^f) = v_{CNN}^f + v_{Tra}^f$;
        \State $Cls(c_n, v^f) = FC(ETIC(c_n) + CMT(v^f))$;
        \State /*\textbf{Backward} propagation:*/
        \State $\theta_{ETIC} = \theta_{ETIC} - $Lr$ \bigtriangledown \lambda L_{MMD}(ETIC(c_n), CMT(v^f))$
        \Statex \qquad \qquad $+ L_{FL}(Cls(c_n, v^f), G)$;
        \State $\theta_{CMT} = \theta_{CMT} - $Lr$ \bigtriangledown \lambda L_{MMD}(ETIC(c_n), CMT(v^f))$ 
        \Statex \qquad \qquad $+ L_{FL}(Cls(c_n, v^f), G)$;
        
    \EndFor
  \end{algorithmic}
\end{breakablealgorithm}

Table 2 provides the numerical study of TASL-Net's architecture. Algorithm 3 summarizes TASL-Net. It integrates features with great diagnostic attention for an accurate computed-aided diagnosis. The optimization of TASL-Net has two parts. The initial step is to reduce the maximum mean discrepancy (MMD) loss \citep{konwer2022temporal} to align the distribution between key TIC information $Z_{TIC}=\{z^x_{TIC}|x=1,2,...,X\}$, with the distribution of dynamic bimodal video representations $Z_{BUS}=\{z^y_{BUS}|y=1,2,...,Y\}$. This alignment is crucial in matching the effective features between the quantified intensity information and the visual expression. Therefore, the optimization of the MMD loss $L_{MMD}$ aims to achieve this effective feature matching:
\begin{equation}
\begin{array}{l}
L_{MMD}=\| \frac{1}{X}\sum\limits_{x=1}^X{z^x_{TIC}}-\frac{1}{Y}\sum\limits_{y=1}^Y{z^y_{BUS}}\|^2,
\end{array}
\end{equation}
where $X=Y=512$ are the numbers of features. Then, to train the TASL-Net, we combine the standard prediction Focal Loss $L_{FL}$ \citep{lin2017focal} and $L_{MMD}$ as the summarized loss to learn the optimal expression of the fused feature:
\begin{equation}
\begin{array}{l}
L=\lambda L_{MMD}+L_{FL}
\end{array}
\end{equation}
where $\lambda=0.83$ is the weight of the MMD loss. This summarized loss is used to train the network and achieve effective feature matching between the quantified intensity information and vision expression, leading to improved classification performance.

\section{Data and Experiments}
\subsection{Datasets}
\begin{table}[!htp]
\centering
\caption{The detailed parameters of lung, breast, and liver datasets.}\label{tab2}
\renewcommand\arraystretch{1.4}
\smallskip\resizebox{0.9\linewidth}{28mm}{\begin{tabular}{c|c|c c|c c|c c}
\hline
\multicolumn{2}{c|}{\multirow{2}{*}{Dataset}} & \multicolumn{2}{c|}{\textbf{Lung}} & \multicolumn{2}{c|}{\textbf{Breast}} & \multicolumn{2}{c}{\textbf{Liver}} \\
\cline{3-8}
\multicolumn{2}{c|}{\multirow{1}{*}{}} & tra.\&val. & tes. & tra.\&val. & tes. & tra.\&val. & tes. \\
\hline
\multirow{3}{*}{\rotatebox{90}{\makecell{Essential \\ Parameter}}} & Total Number & 512 & 51 & 88 & 20 & 100 & 20 \\
\multirow{2}{*}{} & Benign/HCC & 239 & 27 & 34 & 10 & 50 & 10 \\
\multirow{1}{*}{} & Malignant/ICC & 273 & 24 & 54 & 10 & 50 & 10 \\
\hline
\multirow{3}{*}{\rotatebox{90}{\makecell{Collecting \\ Information}}}& Organization & \multicolumn{2}{c|}{Shanghai Pulmonary Hospital} & \multicolumn{2}{c|}{\makecell{Shanghai Jiaotong University \\ School of Medicine \\ Ruijin Hospital}} & \multicolumn{2}{c}{\makecell{Sun-Yat Sen University \\ Cancer Center}} \\
\cline{2-8}
\multirow{2}{*}{} & Scanner & \multicolumn{2}{c|}{GE Logiq E9} & \multicolumn{2}{c|}{MINDRAY Resona 7} & \multicolumn{2}{c}{ACUSON Sequoia 512} \\
\cline{2-8}
\multirow{2}{*}{} & Probe & \multicolumn{2}{c|}{C1-5-D convex probe} & \multicolumn{2}{c|}{L11-3U linear probe} & \multicolumn{2}{c}{4C1 convex probe} \\
\hline
\multirow{3}{*}{\rotatebox{90}{\makecell{Imaging \\ Settings}}} & Mechanical Index & \multicolumn{2}{c|}{0.09-0.16} & \multicolumn{2}{c|}{0.072-0.085} & \multicolumn{2}{c}{0.17-0.19} \\
\multirow{2}{*}{} & Imaging Frequency & \multicolumn{2}{c|}{1-6MHz} & \multicolumn{2}{c|}{3-11MHz} & \multicolumn{2}{c}{1-4MHz} \\
\multirow{1}{*}{} & Imaging Depth & \multicolumn{2}{c|}{5-12cm} & \multicolumn{2}{c|}{3-3.5cm} & \multicolumn{2}{c}{8-12cm} \\
\hline
\end{tabular}}
\end{table}

We evaluated the patient-level-split performance of the proposed TASL-Net network on three challenging datasets, including lung, breast, and liver: 
\begin{itemize}
    \item The \textbf{Lung} dataset consists of 563 bimodal US videos collected from 563 clinical patients of Shanghai Pulmonary Hospital. The videos are generated by LOGIQ E9 US diagnostic system with C1-5-D convex probe. The ratio for training and test is $10:1$.
    \item The \textbf{Breast} dataset consists of 108 bimodal US videos collected from 108 clinical patients of Shanghai Jiaotong University School of Medicine Ruijin Hospital. The videos are generated by MINDRAY Resona 7 US diagnostic system with L11-3U linear probe. The ratio for training and test is $4:1$.
    \item The \textbf{Liver} datasets consists of 120 bimodal US videos collected from 120 clinical patients of Sun-Yat Sen University Cancer Center. The videos are generated by ACUSON Sequoia US diagnostic system with 4C1 convex probe. The ratio for training and test is $5:1$.
\end{itemize}
Table 2 lists the essential parameters, collecting information, and imaging settings of the three datasets in detail. The key-frame selection is efficiently managed by the proposed TIC-based video selector by emulating the temporal attention of sonographers. This selector automatically extracts video clips with rapidly changing perfusion intensity and samples them uniformly. Moreover, this selector is adaptable and can handle US videos of varying lengths, enabling all videos from the three datasets to be processed into 32-frame videos. The frame size of GSUS and CEUS videos are resized to $224\times224$ when feeding into the CMT. Notably, the proposed TASL-Net is designed to mimic realistic clinical scenarios, including variations in video quality, presence of noise, and differences in patient anatomy. To ensure the robustness of our model, we did not use completely clean video data to conduct experiments.

\subsection{Experimental Settings}
We employ the lung data as the main dataset and the breast and liver data as the external validation datasets to demonstrate the superiority of the proposed TASL-Net in three types of experiments.  
\begin{enumerate}
    \item[a.] Ablation study using the lung dataset for classifying benign vs. malignant of lung tumors. The VST Network used in this ablation study of modality, as well as the TASL-Net-derived networks, are trained for 100 epochs with a stochastic gradient descent (SGD) optimizer, batch size of 2, learning rate of $2\times10^{-3}$, weight decay of 0.05, and Focal Loss with parameters (0.2, 4).
    \item[b.] Comparative experiments with SOTA methods using lung dataset for classifying benign vs. malignant of lung tumors. All comparative baseline networks are trained for 100 epochs with a SGD optimizer, batch size of 2, learning rate of $2\times10^{-3}$, weight decay of 0.05, and Focal Loss with parameters (0.2, 4). To ensure a balanced and fair comparison, these comparing benchmark architectures also employ a TIC-based selector for video preprocessing. This strategic choice is aimed at eliminating any potential bias arising from redundant information within the raw videos.
    \item[c.] Generalization study using breast and liver datasets for classifying benign vs. malignant breast tumors, and hepatocellular carcinoma (HCC) vs. intrahepatic cholangiocarcinoma (ICC) of liver tumors. In this part, the TASL-Net is trained for 50 epochs with a SGD optimizer, batch size of 2, learning rate of $2\times10^{-3}$, weight decay of 0.05, and Focal Loss with parameters (0.2, 3).
\end{enumerate}

The TASL-Net is trained for 100 epochs with the SGD optimizer, batch size of 2, and weight decay of 0.05. The parameters of Focal Loss are (0.2, 4). The initial learning rate is set to $2\times10^{-3}$ and decay in a cosine schedule with 2.5 epochs of linear warm-up. All Experiments are based on Python v3.6 and PyTorch v1.7.1 library, and it runs on AMD EPYV 7532 32-Core Processor @ 2.40 GHz with 256 GB RAM and 3 Nvidia GeForce RTX 3090 with CUDA v11.0 and cuDNN v8.1.0.

We adopt standard five-fold-cross validation and testing in each aspect and use four typical evaluation criteria for classification tasks to evaluating the experimental result. 
\begin{enumerate}
    \item[1.] Area Under the Curve (AUC). 
    The AUC measures the area under the Receiver Operating Characteristic (ROC) curve, which plots the True Positive Rate (TPR) against the False Positive Rate (FPR). AUC provides an aggregate measure of a network's classification performance. A higher AUC value indicates better overall performance of the classifier in distinguishing between positive and negative samples 
    \begin{equation}
    \begin{array}{l}
    AUC = \int_0^1 TPR(FPR)d(FPR).
    \end{array}
    \end{equation}
    \item[2.] Classification Accuracy (ACC)
    ACC calculates the ratio of correctly classified samples to the total number of samples. It provides a straightforward measure of overall classification correctness.
    \begin{equation}
    \begin{array}{l}
    ACC = \frac{TP+TN}{TP+TN+FP+FN},
    \end{array}
    \end{equation}
    where TP denotes True Positives, TN denotes True Negatives, FP denotes False Positives, and FN denotes False Negatives.
    \item[3.] Sensitivity (Sens)
    Sensitivity, also known as TPR, measures the proportion of actual positives that are correctly identified by the network. 
    \begin{equation}
    \begin{array}{l}
    Sens = \frac{TP}{TP+FN}.
    \end{array}
    \end{equation}
    \item[4.] Specificity (Spec)
    Specificity calculates the proportion of actual negatives that are correctly identified by the classifier:
    \begin{equation}
    \begin{array}{l}
    Spec = \frac{TN}{TN+FP}.
    \end{array}
    \end{equation}
\end{enumerate}
These metrics collectively provide a comprehensive evaluation of a network's performance, addressing different aspects such as overall accuracy, ability to detect positives (sensitivity), and ability to avoid false alarms (specificity).

\section{Results and Discussion}
\subsection{Accurate Earliest-Enhanced Positions Localization}
\begin{figure}[!htb]
\centering
\centerline{\includegraphics[width=1.0\linewidth]{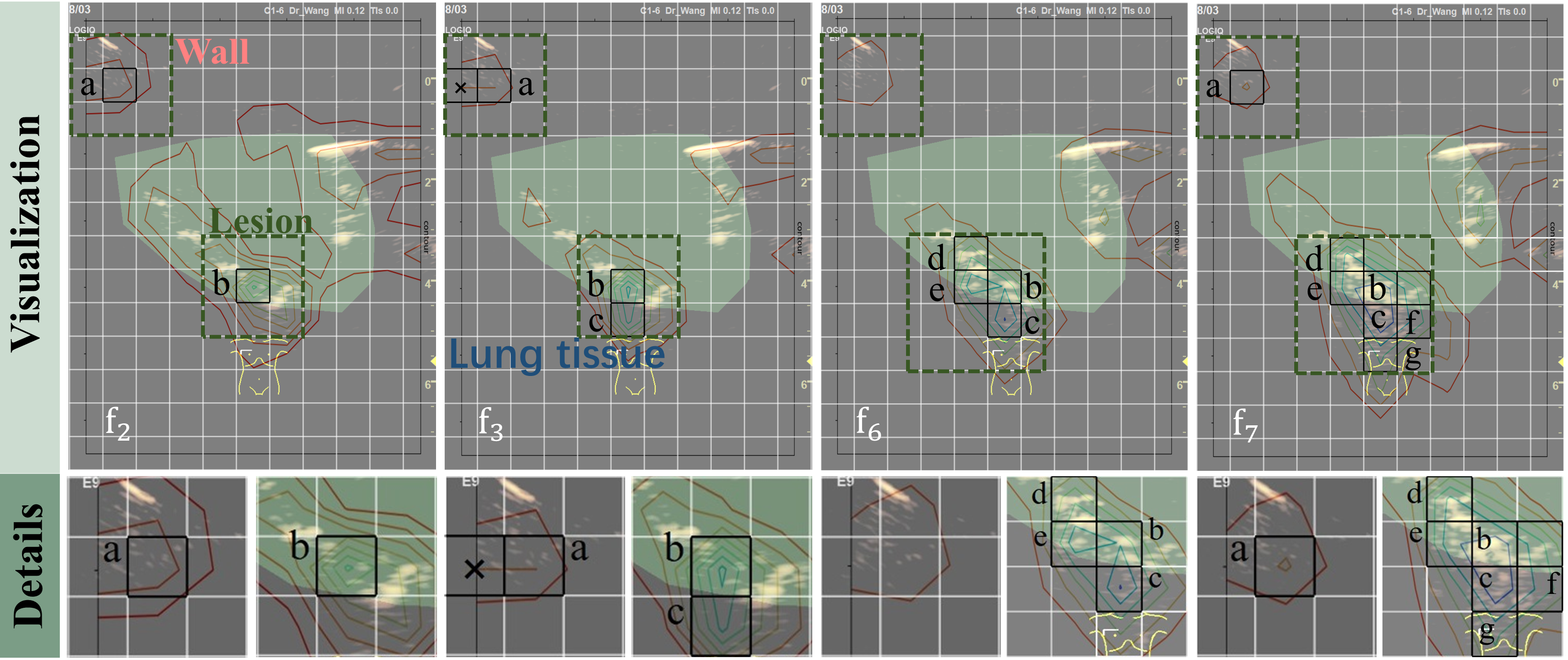}} 
\caption{Take the lung as an example, the TASL-Net has automatically detected the earliest-enhanced positions in the lesion, air-filled lung tissue, and thoracic wall. For better visualization, this figure combines the CEUS image, SSIM contour line, and patch gridding in each sub-figure. The black rectangles represent the earliest-enhanced positions identified by our network. The ``$f_2, f_3, ...$'' denote the ``second, third, ...'' frames in a video. And ``a, b, c,...'' are the positions marker corresponding to Fig. 4.}
\label{fig6}
\end{figure}
A major innovation in the proposed ETIC module is the earliest-enhanced position detector, which accurately analyzes the SSIM changing in the CEUS video and automatically selects the position of great clinical attention. Fig. 4 in Section 3 has provided the SSIM visualization of several adjacent frames. Fig. 6 illustrates four early frames of a lung CEUS video extracted by the TIC-based selector. For better visualization, we merge the CEUS image, SSIM contour line, and patch gridding in each sub-figure. Perfusion inside the tumor begins to enhance at patches b, d, and e from the sixth frame. In the lung tissue, enhancement starts at patches c, f, and g from the seventh frame. The SSIM contour lines exhibit low values at these patches. Hence, given the labels of sonographers as the ground truth (as a measure of evaluation rather than participation in training), the results show that the proposed SSIM-based filter has accurately located the earliest-enhanced positions in the tumor, air-filled lung tissue, and thoracic wall. The accurate localization of the earliest-enhanced position detector has laid the foundation for the TASL-Net to analyze the differences in perfusion variation inside and outside the lesion.

\subsection{Ablation Study}
\begin{table}[t]
\centering
\caption{In the ablation study, the five-fold cross-validation (val.) and testing (tes.) results of the ablation study have demonstrated the effectiveness of tri-attention selective learning.}\label{tab3}
\renewcommand\arraystretch{1.1}
\smallskip\resizebox{0.9\linewidth}{36mm}{\begin{tabular}{c|c|c|c|c|c|c}
\hline
\multicolumn{7}{c}{\textbf{Ablation Study}}\\
\hline
Network & Inputting Mode & stage& AUC & ACC (\%) & Sens (\%) & Spec (\%)\\
\hline
\multirow{8}{*}{\textbf{VST}} & \multirow{2}{*}{$\mathcal{GSUS}$} & val. & 0.73 & 70.70 & 66.95 & 73.99\\
\multirow{7}{*}{} & \multirow{1}{*}{} & tes. & 0.63 & 60.29 & 45.19 & 77.50\\
\multirow{6}{*}{} & \multirow{2}{*}{$\mathcal{CEUS}$} & val. & 0.76 & 74.22 & 70.71 & 77.29\\
\multirow{5}{*}{} & \multirow{1}{*}{} & tes. & 0.68 & 63.92 & 51.11 & 78.33\\
\multirow{4}{*}{} & \multirow{2}{*}{$\mathcal{BUS}$} & val. & 0.82 & 80.66 & 79.41 & 82.05\\
\multirow{3}{*}{} & \multirow{1}{*}{} & tes. & 0.78 & 77.25 & 66.67 & 89.17\\
\multirow{2}{*}{} & \multirow{2}{*}{\textbf{$\mathcal{BUS+CI}$}} & val. & \textbf{0.84} & \textbf{83.20} & \textbf{83.26} & \textbf{83.15}\\
\multirow{1}{*}{} & \multirow{1}{*}{} & tes. & \textbf{0.79} & \textbf{80.39} & \textbf{70.37} & \textbf{91.67}\\
\hline
\multirow{2}{*}{\makecell{Conventional+ETIC+CMT \\ (w/o temporal)}} & \multirow{2}{*}{$\mathcal{BUS+CI}$} & val. & 0.88 & 84.38 & 80.33 & 85.08\\
\multirow{1}{*}{} & \multirow{1}{*}{} & tes. & 0.82 & 81.96 & 76.30 & 88.33\\
\hline
\multirow{2}{*}{\makecell{TIC+CMT \\ (w/o spatial)}} & \multirow{2}{*}{$\mathcal{BUS+CI}$} & val. & 0.89 & 85.35 & 87.45 & 83.52\\
\multirow{1}{*}{} & \multirow{1}{*}{} & tes. & 0.83 & 81.57 & 72.59 & 91.67\\
\hline
\multirow{2}{*}{\makecell{TIC+ETIC+VST \\ (w/o bimodal)}} & \multirow{2}{*}{$\mathcal{BUS+CI}$} & val. & 0.86 & 85.16 & 79.08 & 90.48\\
\multirow{1}{*}{} & \multirow{1}{*}{} & tes. & 0.82 & 80.78 & 71.85 & 90.83\\
\hline
\multirow{2}{*}{\textbf{TASL-Net(ours)}} & \multirow{2}{*}{\textbf{$\mathcal{BUS+CI}$}} & val. & \textbf{0.90} & \textbf{86.72} & \textbf{88.70} & \textbf{84.98}\\
\multirow{1}{*}{} & \multirow{1}{*}{} & tes. & \textbf{0.86} & \textbf{83.53} & \textbf{77.04} & \textbf{90.83}\\
\hline
\end{tabular}}
\end{table}

\subsubsection{Gain of Bi-modality} Since the inputs to the TASL-Net must contain CEUS videos, we use the VST network as the backbone in this part. In classifying lung tumors, we embed valuable clinical information ($\mathcal{CI}$), including gender, age, smoking history, and respiratory disease history, into the classifier to enhance its performance. As shown in Table 3, the bimodal inputting mode ($\mathcal{BUS}$) improves $0.10-0.15$ on AUC and $13.33\%-16.96\%$ on ACC compare to gray-scale inputting mode ($\mathcal{GSUS}$) and contrast-enhanced inputting mode ($\mathcal{CEUS}$). These findings unequivocally demonstrate that, for deep learning networks, utilizing bimodal videos enhances the accuracy of diagnosis more effectively than single-modal videos. Furthermore, the inclusion of $\mathcal{CI}$ achieves a performance gain of $3.14\%$ in ACC, indicating its value in diagnosing benign and malignant lung tumors. Therefore, we adopt $\mathcal{BUS+CI}$ inputting mode in the following experiments on lung datasets.

\subsubsection{Effectiveness of Tri-Attention} As shown in Table 3 and Fig. 7 (a), we derived three networks from the TASL-Net in this part.
\begin{enumerate}
    \item[a.] ``Conventional+ETIC+CMT'', which replaces the TIC-based video selector in TASL-Net with a conventional preprocessing method that uniformly samples the entire video. Comparison between "Conventional+ETIC+CMT" and TASL-Net reveals a decline of 0.04 AUC and $1.57\%$ ACC upon the replacement of TIC-based selector. While two networks extract videos of equal length, this decline can be attributed to the inability of the former to filter out irrelevant diagnostic information. In contrast, our TIC-based selector emulates sonographers' video browsing manner, prioritizing significant video clips. These results show that integrating sonographers' temporal attention in the network effectively reduces the impact of redundant information. 
    \item[b.] ``TIC+CMT'', which uses TIC-based video selector and CMT module for prediction, is a network constructed by removing the ETIC module from TASL-Net. When comparing ``TIC+CMT'' to TASL-Net, the removal of ETIC decreases the AUC of 0.03 and ACC of $1.96\%$. Without the guidance of ETIC module, the former is unable to analyze variations in perfusion enhancement speed among different tissues. It can only capture the general perfusion characteristics in the video and the bi-modal comprehensive features for prediction. These findings underscore the importance of embedding sonographers' spatial attention to highlight crucial perfusion positions for TASL-Net.
    \item[c.] ``TIC+ETIC+VST'', which replaces the CMT module in TASL-Net with VST for prediction. Although this network has focused on high-diagnostic video clips and perfusion enhancement positions leveraging the guidance of temporal and spatial attention, it lacks the capacity to model detailed spatial structural features. However, this capacity is crucial for modeling the textural structural features of tumors and health tissue in GSUS videos. Comparing ``TIC+ETIC+VST'' vs. TASL-Net, the absence of CMT backbone framework has the most pronounced impact on performance, resulting in a decrease of 0.04 in AUC and $2.75\%$ in ACC. This demonstrates the significance of the bimodal attention-guided mutual encoding strategy in enhancing diagnostic accuracy.
\end{enumerate}

\begin{figure}[!htb]
\centering
\centerline{\includegraphics[width=0.8\linewidth]{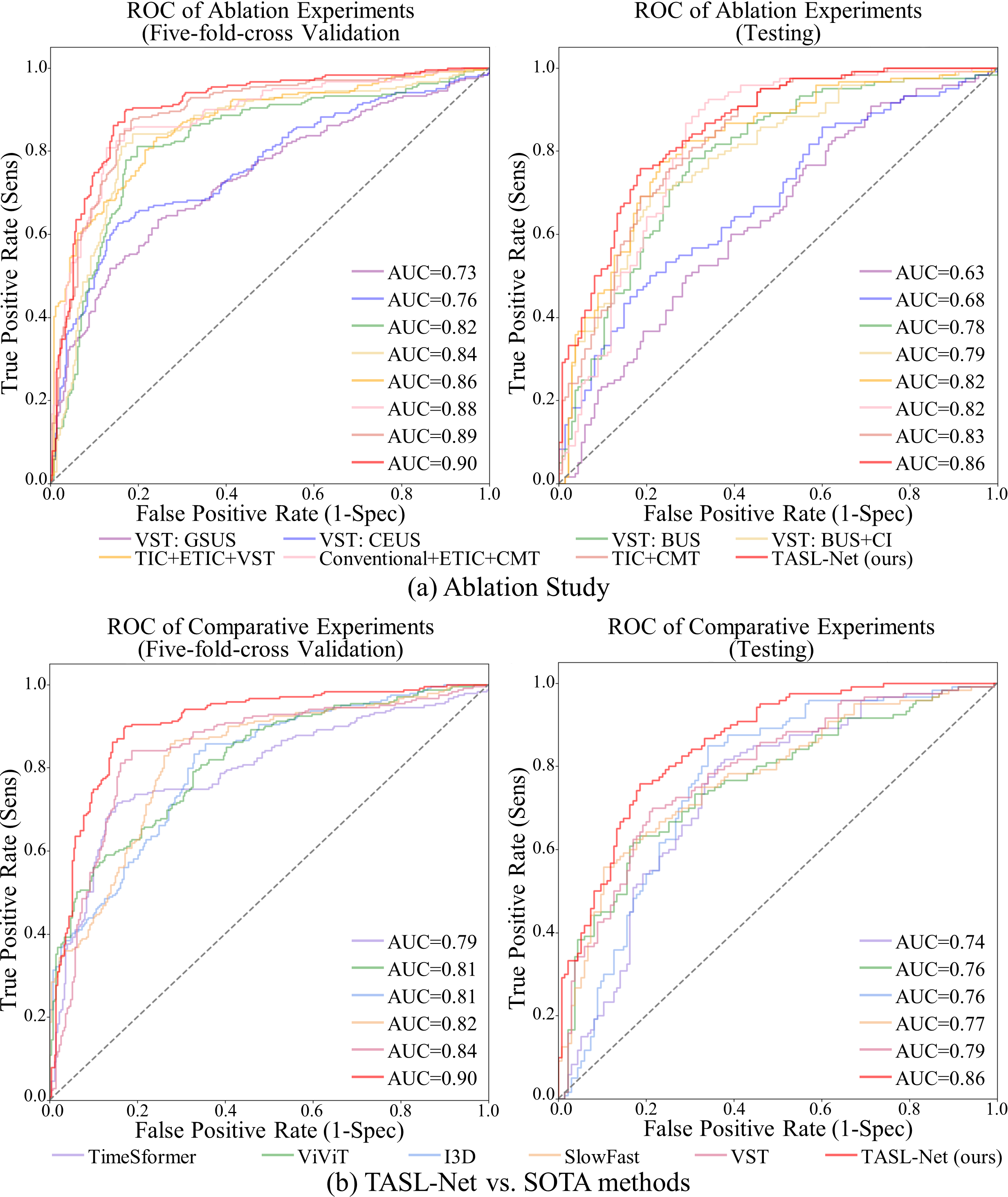}} 
\caption{(a) The ablation study has demonstrated the effectiveness of tri-attention selective learning. (b) TASL-Net has achieved the best ROC compared to SOTA methods.}
\label{fig7}
\end{figure}

\subsection{Comparison with SOTA Methods}
\begin{table}[!htb]
\centering
\caption{The TASL-Net has achieved the highest AUC and ACC compared to five relevant SOTA networks.}\label{tab4}
\renewcommand\arraystretch{1.1}
\smallskip\resizebox{0.9\linewidth}{30mm}{\begin{tabular}{c|c|c|c|c|c|c|c}
\hline
\multicolumn{8}{c}{\textbf{TASL-Net vs. SOTA Methods}}\\
\hline
Type & Network & \#Param(MB) & Stage & AUC & ACC (\%) & Sens (\%) & Spec (\%)\\
\hline
\multirow{6}{*}{Tran-based} & \multirow{2}{*}{TimeSformer} & \multirow{2}{*}{122} & val. & 0.79 & 75.98 & 71.55 & 79.85\\
\multirow{5}{*}{} & \multirow{1}{*}{} & \multirow{1}{*}{} & tes. &0.74 & 72.55 & 68.15 & 77.50\\

\multirow{4}{*}{} & \multirow{2}{*}{ViViT} & \multirow{2}{*}{86} & val. & 0.81 & 78.52 & 74.48 & 82.05\\
\multirow{3}{*}{} & \multirow{1}{*}{} & \multirow{1}{*}{} & tes. & 0.76 & 75.69 & 70.37 & 81.67\\

\multirow{2}{*}{} & \multirow{2}{*}{\textbf{VST}}  & \multirow{2}{*}{\textbf{105}} & val. & \textbf{0.84} & \textbf{83.20} & \textbf{83.26} & \textbf{83.15}\\
\multirow{1}{*}{} & \multirow{1}{*}{}  & \multirow{1}{*}{}  & tes. & \textbf{0.79} & \textbf{80.39} & \textbf{70.37} & \textbf{91.67}\\
\hline
\multirow{4}{*}{Conv-based} & \multirow{2}{*}{I3D} & \multirow{2}{*}{25} & val. & 0.81 & 75.39 & 72.80 & 77.66\\
\multirow{3}{*}{} & \multirow{1}{*}{} & \multirow{1}{*}{}  & tes. & 0.76 & 72.55 & 67.41 & 78.33\\

\multirow{2}{*}{} & \multirow{2}{*}{SlowFast} & \multirow{2}{*}{33} & val. & 0.82 & 77.34 & 72.80 & 81.32\\
\multirow{1}{*}{} & \multirow{1}{*}{} & \multirow{1}{*}{}  & tes. & 0.77 & 75.29 & 70.04 & 80.83\\
\hline
\multirow{2}{*}{Conv-Tran} & \multirow{2}{*}{\textbf{TASL-Net(ours)}} & \multirow{2}{*}{\textbf{147}} & val. & \textbf{0.90} & \textbf{86.72} & \textbf{88.70} & \textbf{84.98}\\
\multirow{1}{*}{} & \multirow{1}{*}{} & \multirow{1}{*}{}  & tes. & \textbf{0.86} & \textbf{83.53} & \textbf{77.04} & \textbf{90.83}\\
\hline
\end{tabular}}
\end{table}

The superiority of TASL-Net in bimodal US video classification has been demonstrated by comparison with five SOTA video classification networks, including three transformer-based networks (TimeSformer, ViViT, and VST) and two convolution-based networks (I3D \citep{carreira2017quo} and SlowFast \citep{feichtenhofer2019slowfast}). As shown in Fig. 7 (b) and Table 4, our network has achieved an increase in AUC of 0.07-0.12 and ACC of $6.06\%$-$11.33\%$ higher than these networks. These results highlight the capability of our TASL-Net agent in combining the advantages of convolution and transformer. This indicates that the embedding of \textit{tri-attention of sonographers} and the \textit{mutual encoding strategy} has cooperatively led to a good performance.

We have provided a comparison of the capacity of TASL-Net compared to other networks. The performance improvement cannot be solely attributed to an increase in network capacity. The results presented in Table 4 has show that the capacity increase results in an average improvement with 0.02 AUC and $3\%$ ACC. In contrast, the results presented in Table 3 demonstrate that tri-attention selective learning leads to an average improvement with 0.07 AUC and $4\%$ ACC. The results of comparative experiments and ablation studies have demonstrated that the primary driver lies in the effective integration of tri-attention.

\subsection{External Validation for Generalization}
\begin{table}[!htb]
 \centering
\caption{The TASL-Net has good performance on the breast and liver datasets.}\label{tab5}
\renewcommand\arraystretch{1.1}
\smallskip\resizebox{0.75\linewidth}{29mm}{\begin{tabular}{c|c|c|c|c|c|c}
\hline
\multicolumn{7}{c}{\textbf{Generalization Study}}\\
\hline
Dataset & network & Stage & AUC & ACC (\%) & Sens (\%) & Spec (\%)\\
\hline
\multirow{6}{*}{Breast} & \multirow{2}{*}{SlowFast} & val. & 0.81 & 74.32 & 71.18 & 76.30\\
\multirow{5}{*}{} & \multirow{1}{*}{} & tes. & 0.77 &  74.00 & 71.00 & 75.00\\
\multirow{4}{*}{} & \multirow{2}{*}{VST} & val. & 0.80 & 76.82 & 69.41 & 79.63\\
\multirow{3}{*}{} & \multirow{1}{*}{} & tes. & 0.76 &  74.00 & 73.00 & 76.00\\
\multirow{2}{*}{} & \textbf{\multirow{2}{*}{TASL-Net(ours)}} & val. & \textbf{0.89} & \textbf{84.09} & \textbf{83.02 }& \textbf{85.71}\\
\multirow{1}{*}{} & \textbf{\multirow{1}{*}{}} & tes. & \textbf{0.86} & \textbf{81.00} & \textbf{78.00} & \textbf{84.00}\\
\hline
\multirow{6}{*}{Liver} & \multirow{2}{*}{SlowFast} & val. & 0.89 & 85.00 & 87.00 & 83.00\\
\multirow{5}{*}{} & \multirow{1}{*}{} & tes. & 0.86 & 84.00 & 85.00 & 82.00\\
\multirow{4}{*}{} & \multirow{2}{*}{VST} & val. & 0.90 & 87.80 & 87.60 & 88.00\\
\multirow{3}{*}{} & \multirow{1}{*}{} & tes. & 0.88 & 86.00 & 84.00 & 87.00\\
\multirow{2}{*}{} & \textbf{\multirow{2}{*}{TASL-Net(ours)}} & val. & \textbf{0.98} & \textbf{94.00} & \textbf{96.00} & \textbf{92.00}\\
\multirow{1}{*}{} & \textbf{\multirow{1}{*}{}} & tes. & \textbf{0.97} & \textbf{92.00} & \textbf{94.00} & \textbf{90.00}\\
\hline
\end{tabular}}
\end{table}
We compare the performances between TASL-Net with two sub-optimal networks, VST and SlowFast, on the two external validation datasets (breast and liver). In this part, no available clinical information is incorporated into the network. The results presented in Table 5 and Fig. 8 have proven the good generalization of TASL-Net in intelligently classifying bimodal US videos of different cancers.

External validation involves direct training on breast or liver data, rather than pre-training on lung data and fine-tuning on breast or liver data. This training approach is chosen with the purpose of assessing the effectiveness of tri-attention learning in scenarios with small datasets. Despite using two smaller-scale datasets, the results remain notably satisfactory. This further underscores the conclusion that integration medical domain knowledge contributes to obtaining robust deep-learning model on small medical datasets \citep{xie2021survey}. In other words, embedding the diagnostic attention of sonographers provides the network with significant and powerful information, i.e., medical domain knowledge, that cannot be extracted from the video alone.

\subsection{Feature Visualization}
\begin{figure}[!htb]
\centering
\centerline{\includegraphics[width=1.0\linewidth]{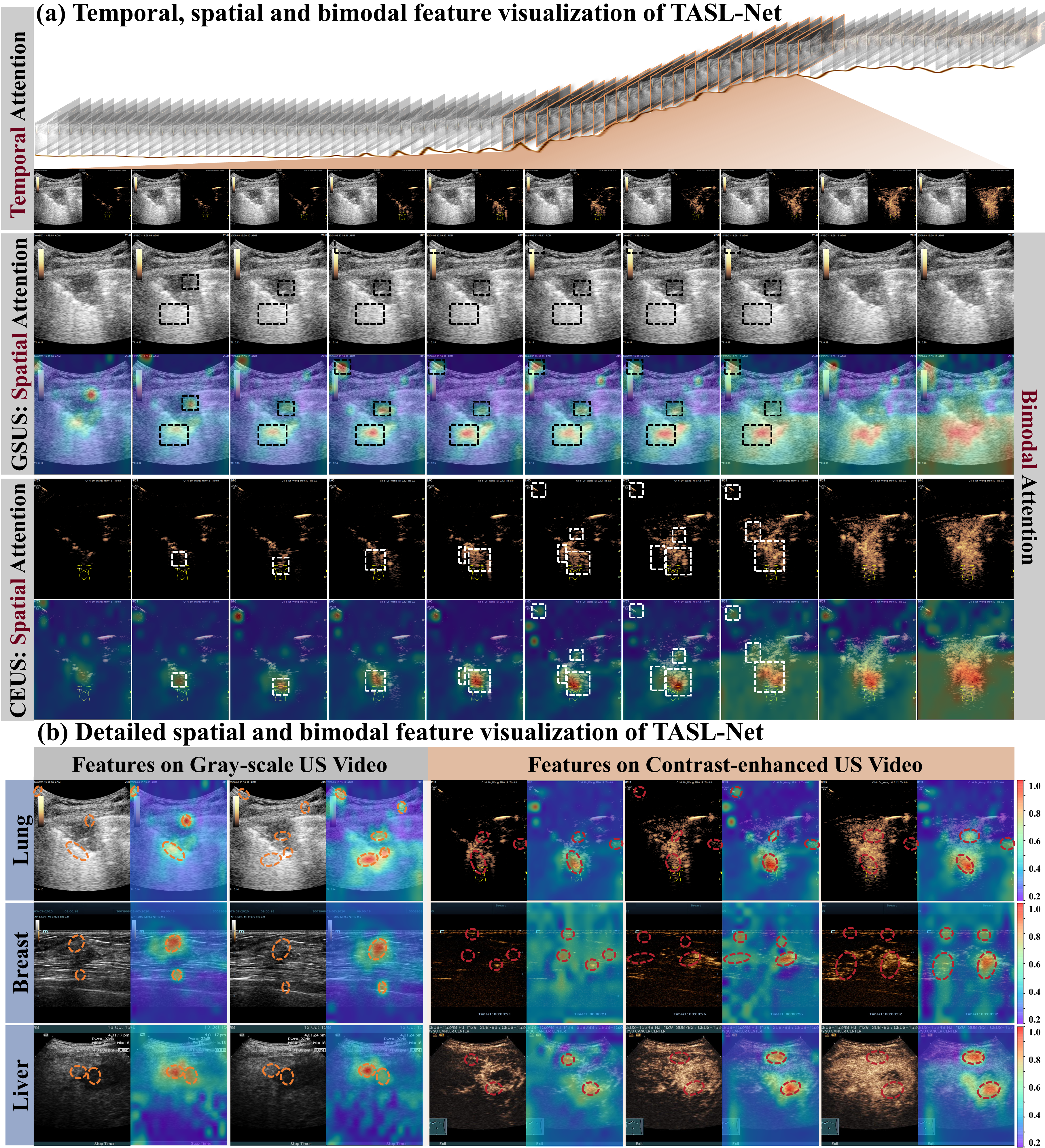}} 
\caption{The TASL-Net has successfully captured texture and structure features in GSUS videos and dynamic perfusion variation in CEUS videos.}
\label{fig8}
\end{figure}
The TASL-Net has successfully captured a comprehensive set of features from both GSUS and CEUS videos, leveraging tri-attention selective learning and mutual encoding strategies. In particular, TASL-Net can extract temporal, spatial, and bimodal features with high accuracy, much like experienced sonographers. Fig. 8 (a) provides an example of feature visualization for a lung video. In the temporal dimension, it focuses on rapidly changing video clips. In the spatial dimension, TASL-Net extracts features from perfusion-enhanced positions on different tissues inside and outside the tumor. In the bimodal dimension, TASL-Net shows different feature emphases in GSUS and CEUS frames. On the GSUS frames, it emphasizes texture and structure features. As shown in the black boxes in Fig. 8 (a), there is no significant change in the position of important features. In contrast, the white boxes demonstrate that with the enhancement of perfusion, the important features in each tissue show a trend of change from nothing to existing and from small to large on the CEUS frames. The change trend of features is consistent with the change trend of perfusion. That is, TASL-Net has effectively extracted the perfusion change information of each tissue in CEUS videos.

For better visualization, we have provided a more comprehensive and detailed explanation in Fig. 8(b). In all three datasets, TASL-Net has shown different types of feature attention on the video frames of GSUS and CEUS modal. As shown in Fig. 9(b), it attends to the texture features of the thoracic wall, internal tumor, tumor boundary, and the lung tissue on the GSUS frames, as indicated by the orange circle. Additionally, TASL-Net focuses on the perfusion differences between tissues inside and outside the tumor, represented by the red circles at various positions in the same frame, similar to the spatial attention of sonographers.

\section{Conclusions}
This paper presents the TASL-Net, an intelligent bimodal ultrasound video classifier leveraging the triple attention of sonographers. Three core modules collectively address a previously unsolved challenge that automatic integration of diagnostic attention with deep learning. Each module makes indispensable technical contributions: a. TIC-based Video Selector: An adaptive inflection-point detection algorithm, which adaptively identifies video clips characterized by dynamic perfusion changes and overcomes patient-specific variations of key video clips. b. ETIC: An adaptive method to identify and analyze TICs at positions emphasized by sonographers, which effectively overcomes patient-specific variations of key positions. c. CMT: A mutual encoding strategy to capture both tissue texture and structure in GSUS videos and dynamic perfusion variations in CEUS videos. 

Extensive ablation and comparison experiments have demonstrated the effectiveness and superiority of the TASL-Net. Two conclusions can be drawn from the results: a. Enhanced Performance: TASL-Net demonstrates improved performance compared to the SOTA methods in terms of AUC and accuracy in lung, breast, and liver datasets. This underscores the robustness and effectiveness of TASL-Net in various diagnostic scenarios. b. Good Generalizability: TASL-Net exhibits good generalization, effectively adapting to different types of US video without significant performance degradation. This indicates its potential for wide application in clinical practice.

Although our datasets are sufficient and of high quality, further research needs to be supported by diverse datasets to lucubrate the clinical generalization and practicality of our network. Therefore, our future efforts will focus on expanding our datasets to include more diverse samples from various medical centers and cancers. More importantly, the significant computation and memory cost associated with bimodal US video analysis remains an intractable challenge. To address this, our future work will concentrate on developing lightweight video analysis networks that maintain high performance while reducing computational requirements. Furthermore, with the numerous breakthroughs of large models in computer vision, we also intend to investigate the performance of existing video large models on bimodal US videos. This exploration will provide insights into potential synergies between large models and our proposed tri-attention selective learning.

\section{Acknowledgments}
This work was supported by the National Natural Science Foundation of China [91959127, 82372096, 82072020, and 82272007] and Shanghai Medical Innovation Research Project [22Y11911400], and Shanghai Technical Standard Project [19DZ2203300].



\bibliographystyle{model5-names}\biboptions{authoryear}
\bibliography{mybibfile}

\begin{thebibliography}{47}
\expandafter\ifx\csname natexlab\endcsname\relax\def\natexlab#1{#1}\fi
\providecommand{\url}[1]{\texttt{#1}}
\providecommand{\href}[2]{#2}
\providecommand{\path}[1]{#1}
\providecommand{\DOIprefix}{doi:}
\providecommand{\ArXivprefix}{arXiv:}
\providecommand{\URLprefix}{URL: }
\providecommand{\Pubmedprefix}{pmid:}
\providecommand{\doi}[1]{\href{http://dx.doi.org/#1}{\path{#1}}}
\providecommand{\Pubmed}[1]{\href{pmid:#1}{\path{#1}}}
\providecommand{\bibinfo}[2]{#2}
\ifx\xfnm\relax \def\xfnm[#1]{\unskip,\space#1}\fi
\bibitem[{Alexander et~al.(2020)Alexander, Patel, Caserta \& Robbin}]{alexander2020thyroid}
\bibinfo{author}{Alexander, L.~F.}, \bibinfo{author}{Patel, N.~J.}, \bibinfo{author}{Caserta, M.~P.}, \& \bibinfo{author}{Robbin, M.~L.} (\bibinfo{year}{2020}).
\newblock \bibinfo{title}{Thyroid ultrasound: diffuse and nodular disease}.
\newblock {\it \bibinfo{journal}{Radiologic Clinics}\/},  {\it \bibinfo{volume}{58}\/}, \bibinfo{pages}{1041--1057}.
\bibitem[{Arnab et~al.(2021)Arnab, Dehghani, Heigold, Sun, Lu{\v{c}}i{\'c} \& Schmid}]{arnab2021vivit}
\bibinfo{author}{Arnab, A.}, \bibinfo{author}{Dehghani, M.}, \bibinfo{author}{Heigold, G.}, \bibinfo{author}{Sun, C.}, \bibinfo{author}{Lu{\v{c}}i{\'c}, M.}, \& \bibinfo{author}{Schmid, C.} (\bibinfo{year}{2021}).
\newblock \bibinfo{title}{Vivit: A video vision transformer}.
\newblock In {\it \bibinfo{booktitle}{Proceedings of the IEEE/CVF International Conference on Computer Vision}\/} (pp. \bibinfo{pages}{6836--6846}).
\bibitem[{Bai et~al.(2018)Bai, Kolter \& Koltun}]{bai2018empirical}
\bibinfo{author}{Bai, S.}, \bibinfo{author}{Kolter, J.~Z.}, \& \bibinfo{author}{Koltun, V.} (\bibinfo{year}{2018}).
\newblock \bibinfo{title}{An empirical evaluation of generic convolutional and recurrent networks for sequence modeling}.
\newblock {\it \bibinfo{journal}{arXiv preprint arXiv:1803.01271}\/}, .
\bibitem[{Bertasius et~al.(2021)Bertasius, Wang \& Torresani}]{bertasius2021space}
\bibinfo{author}{Bertasius, G.}, \bibinfo{author}{Wang, H.}, \& \bibinfo{author}{Torresani, L.} (\bibinfo{year}{2021}).
\newblock \bibinfo{title}{Is space-time attention all you need for video understanding?}
\newblock In {\it \bibinfo{booktitle}{ICML}\/} (p.~\bibinfo{pages}{4}).
\newblock volume~\bibinfo{volume}{2}.
\bibitem[{Bi et~al.(2021)Bi, Zhou, Zhang, Shen, Chen, Cong, Zhu, Tang, Yuan \& Wang}]{bi2021us}
\bibinfo{author}{Bi, K.}, \bibinfo{author}{Zhou, R.-r.}, \bibinfo{author}{Zhang, Y.}, \bibinfo{author}{Shen, M.-j.}, \bibinfo{author}{Chen, H.-w.}, \bibinfo{author}{Cong, Y.}, \bibinfo{author}{Zhu, H.-m.}, \bibinfo{author}{Tang, C.-h.}, \bibinfo{author}{Yuan, J.}, \& \bibinfo{author}{Wang, Y.} (\bibinfo{year}{2021}).
\newblock \bibinfo{title}{Us contrast agent arrival time difference ratio for benign versus malignant subpleural pulmonary lesions}.
\newblock {\it \bibinfo{journal}{Radiology}\/},  {\it \bibinfo{volume}{301}\/}, \bibinfo{pages}{200--210}.
\bibitem[{C{\u{a}}leanu et~al.(2021)C{\u{a}}leanu, S{\^\i}rbu \& Simion}]{cualeanu2021deep}
\bibinfo{author}{C{\u{a}}leanu, C.~D.}, \bibinfo{author}{S{\^\i}rbu, C.~L.}, \& \bibinfo{author}{Simion, G.} (\bibinfo{year}{2021}).
\newblock \bibinfo{title}{Deep neural architectures for contrast enhanced ultrasound (ceus) focal liver lesions automated diagnosis}.
\newblock {\it \bibinfo{journal}{Sensors}\/},  {\it \bibinfo{volume}{21}\/}, \bibinfo{pages}{4126}.
\bibitem[{Caremani et~al.(2008)Caremani, Benci, Lapini, Tacconi, Caremani, Ciccotosto \& Magnolfi}]{caremani2008contrast}
\bibinfo{author}{Caremani, M.}, \bibinfo{author}{Benci, A.}, \bibinfo{author}{Lapini, L.}, \bibinfo{author}{Tacconi, D.}, \bibinfo{author}{Caremani, A.}, \bibinfo{author}{Ciccotosto, C.}, \& \bibinfo{author}{Magnolfi, A.} (\bibinfo{year}{2008}).
\newblock \bibinfo{title}{Contrast enhanced ultrasonography (ceus) in peripheral lung lesions: a study of 60 cases}.
\newblock {\it \bibinfo{journal}{Journal of Ultrasound}\/},  {\it \bibinfo{volume}{11}\/}, \bibinfo{pages}{89--96}.
\bibitem[{Carreira \& Zisserman(2017)}]{carreira2017quo}
\bibinfo{author}{Carreira, J.}, \& \bibinfo{author}{Zisserman, A.} (\bibinfo{year}{2017}).
\newblock \bibinfo{title}{Quo vadis, action recognition? a new model and the kinetics dataset}.
\newblock In {\it \bibinfo{booktitle}{proceedings of the IEEE Conference on Computer Vision and Pattern Recognition}\/} (pp. \bibinfo{pages}{6299--6308}).
\bibitem[{Chen et~al.(2021)Chen, Wang, Niu, Liu, Li \& Gong}]{chen2021domain}
\bibinfo{author}{Chen, C.}, \bibinfo{author}{Wang, Y.}, \bibinfo{author}{Niu, J.}, \bibinfo{author}{Liu, X.}, \bibinfo{author}{Li, Q.}, \& \bibinfo{author}{Gong, X.} (\bibinfo{year}{2021}).
\newblock \bibinfo{title}{Domain knowledge powered deep learning for breast cancer diagnosis based on contrast-enhanced ultrasound videos}.
\newblock {\it \bibinfo{journal}{IEEE Transactions on Medical Imaging}\/},  {\it \bibinfo{volume}{40}\/}, \bibinfo{pages}{2439--2451}.
\bibitem[{Dai et~al.(2021)Dai, Liu, Le \& Tan}]{dai2021coatnet}
\bibinfo{author}{Dai, Z.}, \bibinfo{author}{Liu, H.}, \bibinfo{author}{Le, Q.~V.}, \& \bibinfo{author}{Tan, M.} (\bibinfo{year}{2021}).
\newblock \bibinfo{title}{Coatnet: Marrying convolution and attention for all data sizes}.
\newblock {\it \bibinfo{journal}{Advances in neural information processing systems}\/},  {\it \bibinfo{volume}{34}\/}, \bibinfo{pages}{3965--3977}.
\bibitem[{Du et~al.(2012)Du, Wang, Wan, Hua, Fang, Chen \& Li}]{du2012differentiating}
\bibinfo{author}{Du, J.}, \bibinfo{author}{Wang, L.}, \bibinfo{author}{Wan, C.-F.}, \bibinfo{author}{Hua, J.}, \bibinfo{author}{Fang, H.}, \bibinfo{author}{Chen, J.}, \& \bibinfo{author}{Li, F.-H.} (\bibinfo{year}{2012}).
\newblock \bibinfo{title}{Differentiating benign from malignant solid breast lesions: combined utility of conventional ultrasound and contrast-enhanced ultrasound in comparison with magnetic resonance imaging}.
\newblock {\it \bibinfo{journal}{European journal of radiology}\/},  {\it \bibinfo{volume}{81}\/}, \bibinfo{pages}{3890--3899}.
\bibitem[{Ebadi et~al.(2021)Ebadi, Krishnaswamy, Bolouri, Zonoobi, Greiner, Meuser-Herr, Jaremko, Kapur, Noga \& Punithakumar}]{ebadi2021automated}
\bibinfo{author}{Ebadi, S.~E.}, \bibinfo{author}{Krishnaswamy, D.}, \bibinfo{author}{Bolouri, S. E.~S.}, \bibinfo{author}{Zonoobi, D.}, \bibinfo{author}{Greiner, R.}, \bibinfo{author}{Meuser-Herr, N.}, \bibinfo{author}{Jaremko, J.~L.}, \bibinfo{author}{Kapur, J.}, \bibinfo{author}{Noga, M.}, \& \bibinfo{author}{Punithakumar, K.} (\bibinfo{year}{2021}).
\newblock \bibinfo{title}{Automated detection of pneumonia in lung ultrasound using deep video classification for covid-19}.
\newblock {\it \bibinfo{journal}{Informatics in Medicine Unlocked}\/},  {\it \bibinfo{volume}{25}\/}, \bibinfo{pages}{100687}.
\bibitem[{Feichtenhofer et~al.(2019)Feichtenhofer, Fan, Malik \& He}]{feichtenhofer2019slowfast}
\bibinfo{author}{Feichtenhofer, C.}, \bibinfo{author}{Fan, H.}, \bibinfo{author}{Malik, J.}, \& \bibinfo{author}{He, K.} (\bibinfo{year}{2019}).
\newblock \bibinfo{title}{Slowfast networks for video recognition}.
\newblock In {\it \bibinfo{booktitle}{Proceedings of the IEEE/CVF international conference on computer vision}\/} (pp. \bibinfo{pages}{6202--6211}).
\bibitem[{Feng et~al.(2021)Feng, Song, Chen, Chen, Ni \& Chen}]{feng2021convolutional}
\bibinfo{author}{Feng, X.}, \bibinfo{author}{Song, D.}, \bibinfo{author}{Chen, Y.}, \bibinfo{author}{Chen, Z.}, \bibinfo{author}{Ni, J.}, \& \bibinfo{author}{Chen, H.} (\bibinfo{year}{2021}).
\newblock \bibinfo{title}{Convolutional transformer based dual discriminator generative adversarial networks for video anomaly detection}.
\newblock In {\it \bibinfo{booktitle}{Proceedings of the 29th ACM International Conference on Multimedia}\/} (pp. \bibinfo{pages}{5546--5554}).
\bibitem[{Frank et~al.(2021)Frank, Schipper, Vaturi, Soldati, Smargiassi, Inchingolo, Torri, Perrone, Mento, Demi et~al.}]{frank2021integrating}
\bibinfo{author}{Frank, O.}, \bibinfo{author}{Schipper, N.}, \bibinfo{author}{Vaturi, M.}, \bibinfo{author}{Soldati, G.}, \bibinfo{author}{Smargiassi, A.}, \bibinfo{author}{Inchingolo, R.}, \bibinfo{author}{Torri, E.}, \bibinfo{author}{Perrone, T.}, \bibinfo{author}{Mento, F.}, \bibinfo{author}{Demi, L.} et~al. (\bibinfo{year}{2021}).
\newblock \bibinfo{title}{Integrating domain knowledge into deep networks for lung ultrasound with applications to covid-19}.
\newblock {\it \bibinfo{journal}{IEEE transactions on medical imaging}\/},  {\it \bibinfo{volume}{41}\/}, \bibinfo{pages}{571--581}.
\bibitem[{Girdhar et~al.(2019)Girdhar, Carreira, Doersch \& Zisserman}]{girdhar2019video}
\bibinfo{author}{Girdhar, R.}, \bibinfo{author}{Carreira, J.}, \bibinfo{author}{Doersch, C.}, \& \bibinfo{author}{Zisserman, A.} (\bibinfo{year}{2019}).
\newblock \bibinfo{title}{Video action transformer network}.
\newblock In {\it \bibinfo{booktitle}{Proceedings of the IEEE/CVF conference on computer vision and pattern recognition}\/} (pp. \bibinfo{pages}{244--253}).
\bibitem[{Gong et~al.(2022)Gong, Zhao, Fan, Li, Guo \& Luo}]{gong2022bus}
\bibinfo{author}{Gong, X.}, \bibinfo{author}{Zhao, X.}, \bibinfo{author}{Fan, L.}, \bibinfo{author}{Li, T.}, \bibinfo{author}{Guo, Y.}, \& \bibinfo{author}{Luo, J.} (\bibinfo{year}{2022}).
\newblock \bibinfo{title}{Bus-net: a bimodal ultrasound network for breast cancer diagnosis}.
\newblock {\it \bibinfo{journal}{International Journal of Machine Learning and Cybernetics}\/},  {\it \bibinfo{volume}{13}\/}, \bibinfo{pages}{3311--3328}.
\bibitem[{Jacobsen et~al.(2022)Jacobsen, Pietersen, Nolsoe, Konge, Graumann \& Laursen}]{jacobsen2022clinical}
\bibinfo{author}{Jacobsen, N.}, \bibinfo{author}{Pietersen, P.~I.}, \bibinfo{author}{Nolsoe, C.}, \bibinfo{author}{Konge, L.}, \bibinfo{author}{Graumann, O.}, \& \bibinfo{author}{Laursen, C.~B.} (\bibinfo{year}{2022}).
\newblock \bibinfo{title}{Clinical applications of contrast-enhanced thoracic ultrasound (cetus) compared to standard reference tests: a systematic review}.
\newblock {\it \bibinfo{journal}{Ultraschall in der Medizin-European Journal of Ultrasound}\/},  {\it \bibinfo{volume}{43}\/}, \bibinfo{pages}{72--81}.
\bibitem[{Jung et~al.(2021)Jung, Jung, Stroszczynski \& Wiesinger}]{jung2021quantification}
\bibinfo{author}{Jung, E.~M.}, \bibinfo{author}{Jung, F.}, \bibinfo{author}{Stroszczynski, C.}, \& \bibinfo{author}{Wiesinger, I.} (\bibinfo{year}{2021}).
\newblock \bibinfo{title}{Quantification of dynamic contrast-enhanced ultrasound (ceus) in non-cystic breast lesions using external perfusion software}.
\newblock {\it \bibinfo{journal}{Scientific Reports}\/},  {\it \bibinfo{volume}{11}\/}, \bibinfo{pages}{17677}.
\bibitem[{Kim et~al.(2017)Kim, Noh, Wilson, Kono, Piscaglia, Jang, Lyshchik, Dietrich, Willmann, Vezeridis et~al.}]{kim2017contrast}
\bibinfo{author}{Kim, T.~K.}, \bibinfo{author}{Noh, S.~Y.}, \bibinfo{author}{Wilson, S.~R.}, \bibinfo{author}{Kono, Y.}, \bibinfo{author}{Piscaglia, F.}, \bibinfo{author}{Jang, H.-J.}, \bibinfo{author}{Lyshchik, A.}, \bibinfo{author}{Dietrich, C.~F.}, \bibinfo{author}{Willmann, J.~K.}, \bibinfo{author}{Vezeridis, A.} et~al. (\bibinfo{year}{2017}).
\newblock \bibinfo{title}{Contrast-enhanced ultrasound (ceus) liver imaging reporting and data system (li-rads) 2017--a review of important differences compared to the ct/mri system}.
\newblock {\it \bibinfo{journal}{Clinical and molecular hepatology}\/},  {\it \bibinfo{volume}{23}\/}, \bibinfo{pages}{280}.
\bibitem[{Konwer et~al.(2022)Konwer, Xu, Bae, Chen \& Prasanna}]{konwer2022temporal}
\bibinfo{author}{Konwer, A.}, \bibinfo{author}{Xu, X.}, \bibinfo{author}{Bae, J.}, \bibinfo{author}{Chen, C.}, \& \bibinfo{author}{Prasanna, P.} (\bibinfo{year}{2022}).
\newblock \bibinfo{title}{Temporal context matters: Enhancing single image prediction with disease progression representations}.
\newblock In {\it \bibinfo{booktitle}{Proceedings of the IEEE/CVF Conference on Computer Vision and Pattern Recognition}\/} (pp. \bibinfo{pages}{18824--18835}).
\bibitem[{Lin et~al.(2017)Lin, Goyal, Girshick, He \& Doll{\'a}r}]{lin2017focal}
\bibinfo{author}{Lin, T.-Y.}, \bibinfo{author}{Goyal, P.}, \bibinfo{author}{Girshick, R.}, \bibinfo{author}{He, K.}, \& \bibinfo{author}{Doll{\'a}r, P.} (\bibinfo{year}{2017}).
\newblock \bibinfo{title}{Focal loss for dense object detection}.
\newblock In {\it \bibinfo{booktitle}{Proceedings of the IEEE international conference on computer vision}\/} (pp. \bibinfo{pages}{2980--2988}).
\bibitem[{Liu et~al.(2020)Liu, Luo, Li, Lu, Wu, Sun, Li \& Yang}]{liu2020convtransformer}
\bibinfo{author}{Liu, Z.}, \bibinfo{author}{Luo, S.}, \bibinfo{author}{Li, W.}, \bibinfo{author}{Lu, J.}, \bibinfo{author}{Wu, Y.}, \bibinfo{author}{Sun, S.}, \bibinfo{author}{Li, C.}, \& \bibinfo{author}{Yang, L.} (\bibinfo{year}{2020}).
\newblock \bibinfo{title}{Convtransformer: A convolutional transformer network for video frame synthesis}.
\newblock {\it \bibinfo{journal}{arXiv preprint arXiv:2011.10185}\/}, .
\bibitem[{Liu et~al.(2022)Liu, Ning, Cao, Wei, Zhang, Lin \& Hu}]{liu2022video}
\bibinfo{author}{Liu, Z.}, \bibinfo{author}{Ning, J.}, \bibinfo{author}{Cao, Y.}, \bibinfo{author}{Wei, Y.}, \bibinfo{author}{Zhang, Z.}, \bibinfo{author}{Lin, S.}, \& \bibinfo{author}{Hu, H.} (\bibinfo{year}{2022}).
\newblock \bibinfo{title}{Video swin transformer}.
\newblock In {\it \bibinfo{booktitle}{Proceedings of the IEEE/CVF Conference on Computer Vision and Pattern Recognition}\/} (pp. \bibinfo{pages}{3202--3211}).
\bibitem[{Niu et~al.(2022)Niu, Xiao, Ma, Qin, Li, Zhang, Zhu \& Jiang}]{niu2022value}
\bibinfo{author}{Niu, Z.}, \bibinfo{author}{Xiao, M.}, \bibinfo{author}{Ma, L.}, \bibinfo{author}{Qin, J.}, \bibinfo{author}{Li, W.}, \bibinfo{author}{Zhang, J.}, \bibinfo{author}{Zhu, Q.}, \& \bibinfo{author}{Jiang, Y.} (\bibinfo{year}{2022}).
\newblock \bibinfo{title}{The value of contrast-enhanced ultrasound enhancement patterns for the diagnosis of sentinel lymph node status in breast cancer: systematic review and meta-analysis}.
\newblock {\it \bibinfo{journal}{Quantitative Imaging in Medicine and Surgery}\/},  {\it \bibinfo{volume}{12}\/}, \bibinfo{pages}{936}.
\bibitem[{Peng et~al.(2020)Peng, Lin, Wu, Wan, Zhao, Liang, Ma, Qin, Liu, Li et~al.}]{peng2020ultrasound}
\bibinfo{author}{Peng, Y.}, \bibinfo{author}{Lin, P.}, \bibinfo{author}{Wu, L.}, \bibinfo{author}{Wan, D.}, \bibinfo{author}{Zhao, Y.}, \bibinfo{author}{Liang, L.}, \bibinfo{author}{Ma, X.}, \bibinfo{author}{Qin, H.}, \bibinfo{author}{Liu, Y.}, \bibinfo{author}{Li, X.} et~al. (\bibinfo{year}{2020}).
\newblock \bibinfo{title}{Ultrasound-based radiomics analysis for preoperatively predicting different histopathological subtypes of primary liver cancer}.
\newblock {\it \bibinfo{journal}{Frontiers in oncology}\/},  {\it \bibinfo{volume}{10}\/}, \bibinfo{pages}{1646}.
\bibitem[{Peng et~al.(2021)Peng, Huang, Gu, Xie, Wang, Jiao \& Conformer}]{peng2021local}
\bibinfo{author}{Peng, Z.}, \bibinfo{author}{Huang, W.}, \bibinfo{author}{Gu, S.}, \bibinfo{author}{Xie, L.}, \bibinfo{author}{Wang, Y.}, \bibinfo{author}{Jiao, J.}, \& \bibinfo{author}{Conformer, Q.~Y.} (\bibinfo{year}{2021}).
\newblock \bibinfo{title}{Local features coupling global representations for visual recognition. in 2021 ieee}.
\newblock In {\it \bibinfo{booktitle}{CVF International Conference on Computer Vision, ICCV}\/} (pp. \bibinfo{pages}{357--366}).
\bibitem[{Qian et~al.(2021)Qian, Pei, Zheng, Xie, Yan, Zhang, Han, Gao, Zhang, Zheng et~al.}]{qian2021prospective}
\bibinfo{author}{Qian, X.}, \bibinfo{author}{Pei, J.}, \bibinfo{author}{Zheng, H.}, \bibinfo{author}{Xie, X.}, \bibinfo{author}{Yan, L.}, \bibinfo{author}{Zhang, H.}, \bibinfo{author}{Han, C.}, \bibinfo{author}{Gao, X.}, \bibinfo{author}{Zhang, H.}, \bibinfo{author}{Zheng, W.} et~al. (\bibinfo{year}{2021}).
\newblock \bibinfo{title}{Prospective assessment of breast cancer risk from multimodal multiview ultrasound images via clinically applicable deep learning}.
\newblock {\it \bibinfo{journal}{Nature biomedical engineering}\/},  {\it \bibinfo{volume}{5}\/}, \bibinfo{pages}{522--532}.
\bibitem[{Sandler et~al.(2018)Sandler, Howard, Zhu, Zhmoginov \& Chen}]{sandler2018mobilenetv2}
\bibinfo{author}{Sandler, M.}, \bibinfo{author}{Howard, A.}, \bibinfo{author}{Zhu, M.}, \bibinfo{author}{Zhmoginov, A.}, \& \bibinfo{author}{Chen, L.-C.} (\bibinfo{year}{2018}).
\newblock \bibinfo{title}{Mobilenetv2: Inverted residuals and linear bottlenecks}.
\newblock In {\it \bibinfo{booktitle}{Proceedings of the IEEE conference on computer vision and pattern recognition}\/} (pp. \bibinfo{pages}{4510--4520}).
\bibitem[{Sartori et~al.(2013)Sartori, Postorivo, Di~Vece, Ermili, Tassinari \& Tombesi}]{sartori2013contrast}
\bibinfo{author}{Sartori, S.}, \bibinfo{author}{Postorivo, S.}, \bibinfo{author}{Di~Vece, F.}, \bibinfo{author}{Ermili, F.}, \bibinfo{author}{Tassinari, D.}, \& \bibinfo{author}{Tombesi, P.} (\bibinfo{year}{2013}).
\newblock \bibinfo{title}{Contrast-enhanced ultrasonography in peripheral lung consolidations: what’s its actual role?}
\newblock {\it \bibinfo{journal}{World Journal of Radiology}\/},  {\it \bibinfo{volume}{5}\/}, \bibinfo{pages}{372}.
\bibitem[{Schafer(2011)}]{schafer2011savitzky}
\bibinfo{author}{Schafer, R.~W.} (\bibinfo{year}{2011}).
\newblock \bibinfo{title}{What is a savitzky-golay filter?[lecture notes]}.
\newblock {\it \bibinfo{journal}{IEEE Signal processing magazine}\/},  {\it \bibinfo{volume}{28}\/}, \bibinfo{pages}{111--117}.
\bibitem[{Schmiedt et~al.(2022)Schmiedt, Simion \& C{\u{a}}leanu}]{schmiedt2022preliminary}
\bibinfo{author}{Schmiedt, K.}, \bibinfo{author}{Simion, G.}, \& \bibinfo{author}{C{\u{a}}leanu, C.~D.} (\bibinfo{year}{2022}).
\newblock \bibinfo{title}{Preliminary results on contrast enhanced ultrasound video stream diagnosis using deep neural architectures}.
\newblock In {\it \bibinfo{booktitle}{2022 International Symposium on Electronics and Telecommunications (ISETC)}\/} (pp. \bibinfo{pages}{1--4}).
\newblock \bibinfo{organization}{IEEE}.
\bibitem[{Schwarz et~al.(2021)Schwarz, Clevert, Ingrisch, Geyer, Schwarze, R{\"u}benthaler \& Armbruster}]{schwarz2021quantitative}
\bibinfo{author}{Schwarz, S.}, \bibinfo{author}{Clevert, D.-A.}, \bibinfo{author}{Ingrisch, M.}, \bibinfo{author}{Geyer, T.}, \bibinfo{author}{Schwarze, V.}, \bibinfo{author}{R{\"u}benthaler, J.}, \& \bibinfo{author}{Armbruster, M.} (\bibinfo{year}{2021}).
\newblock \bibinfo{title}{Quantitative analysis of the time--intensity curve of contrast-enhanced ultrasound of the liver: Differentiation of benign and malignant liver lesions}.
\newblock {\it \bibinfo{journal}{Diagnostics}\/},  {\it \bibinfo{volume}{11}\/}, \bibinfo{pages}{1244}.
\bibitem[{Selva et~al.(2022)Selva, Johansen, Escalera, Nasrollahi, Moeslund \& Clap{\'e}s}]{selva2022video}
\bibinfo{author}{Selva, J.}, \bibinfo{author}{Johansen, A.~S.}, \bibinfo{author}{Escalera, S.}, \bibinfo{author}{Nasrollahi, K.}, \bibinfo{author}{Moeslund, T.~B.}, \& \bibinfo{author}{Clap{\'e}s, A.} (\bibinfo{year}{2022}).
\newblock \bibinfo{title}{Video transformers: A survey}.
\newblock {\it \bibinfo{journal}{arXiv preprint arXiv:2201.05991}\/}, .
\bibitem[{Sidhu et~al.(2018)Sidhu, Cantisani, Dietrich, Gilja, Saftoiu, Bartels, Bertolotto, Calliada, Clevert, Cosgrove et~al.}]{sidhu2018efsumb}
\bibinfo{author}{Sidhu, P.~S.}, \bibinfo{author}{Cantisani, V.}, \bibinfo{author}{Dietrich, C.~F.}, \bibinfo{author}{Gilja, O.~H.}, \bibinfo{author}{Saftoiu, A.}, \bibinfo{author}{Bartels, E.}, \bibinfo{author}{Bertolotto, M.}, \bibinfo{author}{Calliada, F.}, \bibinfo{author}{Clevert, D.-A.}, \bibinfo{author}{Cosgrove, D.} et~al. (\bibinfo{year}{2018}).
\newblock \bibinfo{title}{The efsumb guidelines and recommendations for the clinical practice of contrast-enhanced ultrasound (ceus) in non-hepatic applications: update 2017 (long version)}.
\newblock {\it \bibinfo{journal}{Ultraschall in der Medizin-European journal of ultrasound}\/},  {\it \bibinfo{volume}{39}\/}, \bibinfo{pages}{e2--e44}.
\bibitem[{Soldati et~al.(2019)Soldati, Demi, Smargiassi, Inchingolo \& Demi}]{soldati2019role}
\bibinfo{author}{Soldati, G.}, \bibinfo{author}{Demi, M.}, \bibinfo{author}{Smargiassi, A.}, \bibinfo{author}{Inchingolo, R.}, \& \bibinfo{author}{Demi, L.} (\bibinfo{year}{2019}).
\newblock \bibinfo{title}{The role of ultrasound lung artifacts in the diagnosis of respiratory diseases}.
\newblock {\it \bibinfo{journal}{Expert review of respiratory medicine}\/},  {\it \bibinfo{volume}{13}\/}, \bibinfo{pages}{163--172}.
\bibitem[{Sood et~al.(2019)Sood, Rositch, Shakoor, Ambinder, Pool, Pollack, Mollura, Mullen \& Harvey}]{sood2019ultrasound}
\bibinfo{author}{Sood, R.}, \bibinfo{author}{Rositch, A.~F.}, \bibinfo{author}{Shakoor, D.}, \bibinfo{author}{Ambinder, E.}, \bibinfo{author}{Pool, K.-L.}, \bibinfo{author}{Pollack, E.}, \bibinfo{author}{Mollura, D.~J.}, \bibinfo{author}{Mullen, L.~A.}, \& \bibinfo{author}{Harvey, S.~C.} (\bibinfo{year}{2019}).
\newblock \bibinfo{title}{Ultrasound for breast cancer detection globally: a systematic review and meta-analysis}.
\newblock {\it \bibinfo{journal}{Journal of global oncology}\/},  {\it \bibinfo{volume}{5}\/}, \bibinfo{pages}{1--17}.
\bibitem[{Sperandeo et~al.(2014)Sperandeo, Rotondo, Guglielmi, Catalano, Feragalli \& Trovato}]{sperandeo2014transthoracic}
\bibinfo{author}{Sperandeo, M.}, \bibinfo{author}{Rotondo, A.}, \bibinfo{author}{Guglielmi, G.}, \bibinfo{author}{Catalano, D.}, \bibinfo{author}{Feragalli, B.}, \& \bibinfo{author}{Trovato, G.~M.} (\bibinfo{year}{2014}).
\newblock \bibinfo{title}{Transthoracic ultrasound in the assessment of pleural and pulmonary diseases: use and limitations}.
\newblock {\it \bibinfo{journal}{La radiologia medica}\/},  {\it \bibinfo{volume}{119}\/}, \bibinfo{pages}{729--740}.
\bibitem[{Trimboli et~al.(2020)Trimboli, Castellana, Virili, Havre, Bini, Marinozzi, D’Ambrosio, Giorgino, Giovanella, Prosch et~al.}]{trimboli2020performance}
\bibinfo{author}{Trimboli, P.}, \bibinfo{author}{Castellana, M.}, \bibinfo{author}{Virili, C.}, \bibinfo{author}{Havre, R.~F.}, \bibinfo{author}{Bini, F.}, \bibinfo{author}{Marinozzi, F.}, \bibinfo{author}{D’Ambrosio, F.}, \bibinfo{author}{Giorgino, F.}, \bibinfo{author}{Giovanella, L.}, \bibinfo{author}{Prosch, H.} et~al. (\bibinfo{year}{2020}).
\newblock \bibinfo{title}{Performance of contrast-enhanced ultrasound (ceus) in assessing thyroid nodules: a systematic review and meta-analysis using histological standard of reference}.
\newblock {\it \bibinfo{journal}{La radiologia medica}\/},  {\it \bibinfo{volume}{125}\/}, \bibinfo{pages}{406--415}.
\bibitem[{Wang et~al.(2004)Wang, Bovik, Sheikh \& Simoncelli}]{wang2004image}
\bibinfo{author}{Wang, Z.}, \bibinfo{author}{Bovik, A.~C.}, \bibinfo{author}{Sheikh, H.~R.}, \& \bibinfo{author}{Simoncelli, E.~P.} (\bibinfo{year}{2004}).
\newblock \bibinfo{title}{Image quality assessment: from error visibility to structural similarity}.
\newblock {\it \bibinfo{journal}{IEEE transactions on image processing}\/},  {\it \bibinfo{volume}{13}\/}, \bibinfo{pages}{600--612}.
\bibitem[{Wu et~al.(2021)Wu, Xiao, Codella, Liu, Dai, Yuan \& Zhang}]{wu2021cvt}
\bibinfo{author}{Wu, H.}, \bibinfo{author}{Xiao, B.}, \bibinfo{author}{Codella, N.}, \bibinfo{author}{Liu, M.}, \bibinfo{author}{Dai, X.}, \bibinfo{author}{Yuan, L.}, \& \bibinfo{author}{Zhang, L.} (\bibinfo{year}{2021}).
\newblock \bibinfo{title}{Cvt: Introducing convolutions to vision transformers}.
\newblock In {\it \bibinfo{booktitle}{Proceedings of the IEEE/CVF international conference on computer vision}\/} (pp. \bibinfo{pages}{22--31}).
\bibitem[{Xie et~al.(2021)Xie, Niu, Liu, Chen, Tang \& Yu}]{xie2021survey}
\bibinfo{author}{Xie, X.}, \bibinfo{author}{Niu, J.}, \bibinfo{author}{Liu, X.}, \bibinfo{author}{Chen, Z.}, \bibinfo{author}{Tang, S.}, \& \bibinfo{author}{Yu, S.} (\bibinfo{year}{2021}).
\newblock \bibinfo{title}{A survey on incorporating domain knowledge into deep learning for medical image analysis}.
\newblock {\it \bibinfo{journal}{Medical Image Analysis}\/},  {\it \bibinfo{volume}{69}\/}, \bibinfo{pages}{101985}.
\bibitem[{Xu et~al.(2020)Xu, Ding, Wang, Fu, Zhu, Wang \& Lin}]{xu2020savitzky}
\bibinfo{author}{Xu, Z.-t.}, \bibinfo{author}{Ding, H.}, \bibinfo{author}{Wang, B.-g.}, \bibinfo{author}{Fu, T.-t.}, \bibinfo{author}{Zhu, Y.-l.}, \bibinfo{author}{Wang, W.-p.}, \& \bibinfo{author}{Lin, F.} (\bibinfo{year}{2020}).
\newblock \bibinfo{title}{Savitzky-golay filter based quantitative dynamic contrast-enhanced ultrasound on assessing therapeutic response in mice with hepatocellular carcinoma}.
\newblock {\it \bibinfo{journal}{Journal of Signal Processing Systems}\/},  {\it \bibinfo{volume}{92}\/}, \bibinfo{pages}{315--323}.
\bibitem[{Yang et~al.(2015)Yang, Chen, Wu, Dai \& Fan}]{yang2015effects}
\bibinfo{author}{Yang, W.}, \bibinfo{author}{Chen, M.-H.}, \bibinfo{author}{Wu, W.}, \bibinfo{author}{Dai, Y.}, \& \bibinfo{author}{Fan, Z.-H.} (\bibinfo{year}{2015}).
\newblock \bibinfo{title}{Effects of gray-scale ultrasonography immediate post-contrast on characterization of focal liver lesions}.
\newblock {\it \bibinfo{journal}{BioMed Research International}\/},  {\it \bibinfo{volume}{2015}\/}, \bibinfo{pages}{193178}.
\bibitem[{Yang et~al.(2021)Yang, Dong, Du, Qiang, Wu, Zhao, Yang \& Zia}]{yang2021integrate}
\bibinfo{author}{Yang, W.}, \bibinfo{author}{Dong, Y.}, \bibinfo{author}{Du, Q.}, \bibinfo{author}{Qiang, Y.}, \bibinfo{author}{Wu, K.}, \bibinfo{author}{Zhao, J.}, \bibinfo{author}{Yang, X.}, \& \bibinfo{author}{Zia, M.~B.} (\bibinfo{year}{2021}).
\newblock \bibinfo{title}{Integrate domain knowledge in training multi-task cascade deep learning model for benign--malignant thyroid nodule classification on ultrasound images}.
\newblock {\it \bibinfo{journal}{Engineering Applications of Artificial Intelligence}\/},  {\it \bibinfo{volume}{98}\/}, \bibinfo{pages}{104064}.
\bibitem[{Yang et~al.(2020)Yang, Gong, Guo \& Liu}]{yang2020temporal}
\bibinfo{author}{Yang, Z.}, \bibinfo{author}{Gong, X.}, \bibinfo{author}{Guo, Y.}, \& \bibinfo{author}{Liu, W.} (\bibinfo{year}{2020}).
\newblock \bibinfo{title}{A temporal sequence dual-branch network for classifying hybrid ultrasound data of breast cancer}.
\newblock {\it \bibinfo{journal}{IEEE Access}\/},  {\it \bibinfo{volume}{8}\/}, \bibinfo{pages}{82688--82699}.
\bibitem[{Zhou et~al.(2022)Zhou, Zhang, Du, Jiang, Zhao, Sun, Li, Wan, Wang, Hou et~al.}]{zhou2022contrast}
\bibinfo{author}{Zhou, H.}, \bibinfo{author}{Zhang, C.}, \bibinfo{author}{Du, L.}, \bibinfo{author}{Jiang, J.}, \bibinfo{author}{Zhao, Q.}, \bibinfo{author}{Sun, J.}, \bibinfo{author}{Li, Q.}, \bibinfo{author}{Wan, M.}, \bibinfo{author}{Wang, X.}, \bibinfo{author}{Hou, X.} et~al. (\bibinfo{year}{2022}).
\newblock \bibinfo{title}{Contrast-enhanced ultrasound liver imaging reporting and data system in diagnosing hepatocellular carcinoma: diagnostic performance and interobserver agreement}.
\newblock {\it \bibinfo{journal}{Ultraschall in der Medizin-European Journal of Ultrasound}\/},  {\it \bibinfo{volume}{43}\/}, \bibinfo{pages}{64--71}.

\end{thebibliography}





\end{document}